\documentclass[10pt,twocolumn,letterpaper]{article}

\usepackage{cvpr}
\usepackage{times}
\usepackage{epsfig}
\usepackage{graphicx}
\usepackage{amsmath}
\usepackage{amssymb}
\usepackage{multirow}

\cvprfinalcopy %

\def\cvprPaperID{****} %

\usepackage{amsmath}
\usepackage{amssymb}
\usepackage{array}
\usepackage{subcaption}
\usepackage{float}
\DeclareMathOperator*{\E}{\mathbb{E}}
\usepackage{booktabs}      %
\usepackage[numbers]{natbib}
\usepackage{placeins}
\usepackage{enumitem}

\begin{document}

\title{\textit{S2cGAN}: Semi-Supervised Training of Conditional GANs with Fewer Labels}

\author{Arunava Chakraborty \\
Microsoft Research India \\
\and
Rahul Ragesh \\
Microsoft Research India \\
\and
Mahir Shah \\
Microsoft Research India \\
\and
Nipun Kwatra \\
Microsoft Research India \\
}

\twocolumn[{%
\renewcommand\twocolumn[1][]{#1}%
\maketitle
\begin{center}
    \centering
    \vspace{-12pt}
    \begin{tabular}{c @{\hspace{.5\tabcolsep}}c}
     \includegraphics[width=.8366\textwidth]{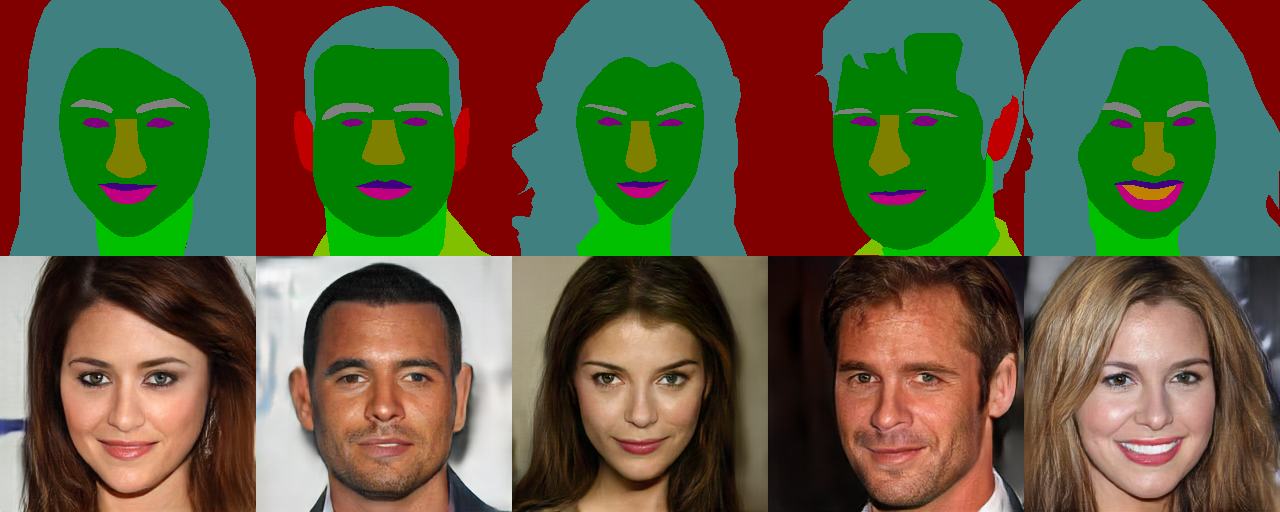} &
     \includegraphics[width=.1333\textwidth]{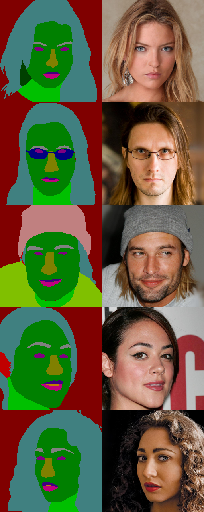} \\
     {\scriptsize Synthesized Images} & {\scriptsize Labeled Pairs}
   \end{tabular}
    \captionof{figure}{We propose a framework for semi-supervised training of conditional GANs, which uses much fewer labels than traditionally required. Here we train a semantic image synthesis network using our framework with just 5 labeled pairs (shown on the right), and around $29000$ unpaired images. Synthesized images and corresponding input semantic maps from the test set are shown on the left. Even with just 5 labelled pairs, the network is able to synthesize high quality results, while accurately respecting the semantic layout.}
    \label{fig:showcase_5_paired_results}
\end{center}%
}]

\newcommand{\sscgan}{\textit{S2cGAN}}
\newcommand{\camhq}{CelebAMask-HQ}
\newcommand{\bltwo}{na\"ive baseline}
\newcommand{\bltwocap}{Na\"ive baseline}
\newcommand{\bltwocapshort}{Na\"ive BL}

\newcommand{\xb}{\textbf{x}}
\newcommand{\Xb}{\textbf{X}}
\newcommand{\px}{p_\Xb}

\newcommand{\zb}{\textbf{z}}
\newcommand{\Zb}{\textbf{Z}}
\newcommand{\pz}{p_\Zb}

\newcommand{\cb}{\textbf{c}}
\newcommand{\Cb}{\textbf{C}}
\newcommand{\pc}{p_\Cb}

\newcommand{\pxc}{p_{\Xb,\Cb}}

\newcommand{\Lb}{\textbf{L}}
\newcommand{\pL}{p_{\Lb}}

\newcommand{\pxL}{p_{\Xb,\Lb}}

\newcommand{\Gb}{\textbf{G}}
\newcommand{\pg}{p_\Gb}
\newcommand{\pgc}{p_{\Gb,\Cb}}

\newcommand{\Sset}{\textbf{S}}
\newcommand{\Uset}{\textbf{U}}

\newcommand{\Ls}{L^{*}}
\newcommand{\Gs}{G^{*}}
\newcommand{\Lbs}{\Lb^{*}}
\newcommand{\Gbs}{\Gb^{*}}

\newcommand{\pgs}{p_{\Gbs}}
\newcommand{\pLs}{p_{\Lbs}}

\begin{abstract}
\vspace{-4pt}
Generative adversarial networks (GANs) have been remarkably successful in learning complex high dimensional real word distributions and generating realistic samples. However, they provide limited control over the generation process. Conditional GANs (cGANs) provide a mechanism to control the generation process by conditioning the output on a user defined input. Although training GANs requires only unsupervised data, training cGANs requires labelled data which can be very expensive to obtain. We propose a framework for semi-supervised training of cGANs which utilizes sparse labels to learn the conditional mapping, and at the same time leverages a large amount of unsupervised data to learn the unconditional distribution. We demonstrate effectiveness of our method on multiple datasets and different conditional tasks.
\end{abstract}

\section{Introduction}
\label{sec:introduction}
GANs have been remarkably successful in generating high dimensional real world data distributions. However, they provide no control in determining the generated output. cGANs help provide this control by conditioning the generated output on conditions such as object classes (e.g. dog, cat, car, etc.) or semantic maps (e.g. pixel level information indicating presence of road, building, tree, etc.). For most real world applications GANs require a lot of training data because of the complexity and high dimensionality of typical data distributions. In the conditional-GAN setting there is the additional requirement of conditioning the output on the input condition, which requires training data labelled with conditional information. Such labelled data can unfortunately be very expensive to obtain, especially for fine grained labels such as semantic maps.

cGANs have two high level tasks -- 1) model the underlying data distribution (typically high dimensional and complex for real world tasks) and 2) learn a way to incorporate conditional mapping during synthesis. Although we don't go into the mechanisms of how a cGAN may be learning this, we note that learning the conditional mapping is in principle a much simpler problem than learning the underlying data distribution. Thus, we reason that it should be possible to learn the conditional mapping task (which requires supervised labels) from much fewer training data, as compared to the large amount of data required for the task of learning the data distribution (which requires only unsupervised data). With this in mind, we develop our semi-supervised method for training cGANs, which utilizes only sparse labels to learn the conditional mapping, and at the same time leverages a large amount of unsupervised data to learn the complex unconditional data distribution. We call our framework the \sscgan{} for \textit{S}emi-\textit{S}upervised \textit{C}onditional GAN. \sscgan{} is able to synthesize high quality results even when trained on very sparse labelled data (see Figure~\ref{fig:showcase_5_paired_results}).

The key insight of our method is a way to incorporate unsupervised examples in the training of conditional GANs. We do this by including an additional labeller network which generates conditional labels from unsupervised inputs. The labeller network is trained jointly with the cGAN. We also propose an unsupervised GAN objective, which combined with the supervised objective achieves the dual purpose of both learning the underlying distribution, as well as learning the conditional mapping. Our method is general and works for any type of conditional GAN, unlike methods such as \cite{lucic2019highfidfewerlabels} which only work for class conditional synthesis. Moreover, our method is very simple to implement, e.g. for semantic image synthesis, we needed only $\approx250$ extra lines of code. The main contributions of our work are:
\begin{enumerate}[nosep]
\item A simple, yet effective technique to incorporate unsupervised samples in training of cGANs.
\item A general framework which works on any type of conditional GAN.
\item Validation that even difficult tasks like semantic image synthesis can be trained with very few labelled pairs.
\end{enumerate}

\section{Background}
\label{sec:background}
We first give a brief background of GANs and cGANs, before describing our method in section~\ref{sec:sscgan}. A GAN consists of a generator and a discriminator. The task of the generator is to map input noise, sampled from a prior $\pz$, to points in the data domain via a mapping $G(\zb; \theta_g)$, where $G$ is the function represented by the generator neural network parameterized by $\theta_g$. The task of the discriminator, on the other hand, is to discriminate between real and generated samples. The discriminator network, parameterized by $\theta_d$, represents a scalar function $D(\xb; \theta_d)$, which typically indicates the predicted probability of the input coming from real vs generated distributions \cite{goodfellow2014generative} \footnote{In the Wasserstein GAN formulation (\cite{arjovsky2017wasserstein}), the discriminator is called a critic and plays a slightly different role. It predicts the Wasserstein distance between the real and fake distributions when maximized under some constraints (also see \cite{arjovsky2017towardsprincipled}). Very interestingly the minimax setting remains.}. The generator and discriminator then play a game, where the generator tries to generate examples to fool the discriminator, while the discriminator tries to improve its discriminative power. More formally, G and D play the following minimax game
\begin{align}
\label{eq:gan_minimax}
\min_{G} \max_{D} V(D, G),
\end{align}
where
\begin{align}
\label{eq:gan_objective}
V(D, G) := \E_{\xb\sim \px}[log D(\xb)] + \E_{\zb\sim \pz}[log(1 - D(G(\zb)))].
\end{align}
Here $\px$ is the real underlying data distribution, while $\pz$ is some noise distribution (typically Gaussian or uniform).

In the case of conditional GANs (cGANs), the generator takes as input a conditional (and optionally a noise sample) to generate a fake sample. The discriminator now takes as input both a data sample and the corresponding conditional, and predicts the probability of the input pair coming from real vs generated distribution. The formulation stays similar, where $G$ and $D$ play the minimax game on the following objective:
\begin{multline}
\label{eq:cgan_objective}
V_c(D, G) := \\
\E_{(\xb,\cb)\sim \pxc}[log D(\xb, \cb)] + \E_{\cb\sim \pc}[log(1 - D(G(\cb), \cb))].
\end{multline}
For simplicity, we have ignored noise input in the above equation. Here $\pxc$ is the joint probability distribution of the real data and the corresponding conditionals, while $\pc$ is the probability distribution of the conditionals.

For the purpose of this paper we will assume that at the end of the minimax optimization, the two distributions being discriminated by the discriminator converge. In the case of vanilla GAN this means that the distribution $\pg(\xb)=\E_{\pz}[\pg(\xb|\zb)]$ induced by the generator matches the real probability distribution $\px$. Here $\pg(\xb|\zb) = \delta(\xb-G(\zb))$ denotes the conditional distribution induced by the deterministic generator. Similarly, in the case of cGAN, this means that the real joint distribution $\pxc$ matches the joint distribution $\pgc(\xb,\cb)=\pg(\xb|\cb)\pc(\cb)$ induced by the generator. Again, $\pg(\xb|\cb) = \delta(\xb-G(\cb))$ denotes the conditional distribution induced by the deterministic generator. See \cite{arjovsky2017towardsprincipled} for conditions when this assumption is reasonable.

\section{Semi-Supervised Conditional GAN}
\label{sec:sscgan}

\begin{figure*}[h]
\vspace{-24pt}
\centering
  \includegraphics[width=\textwidth]{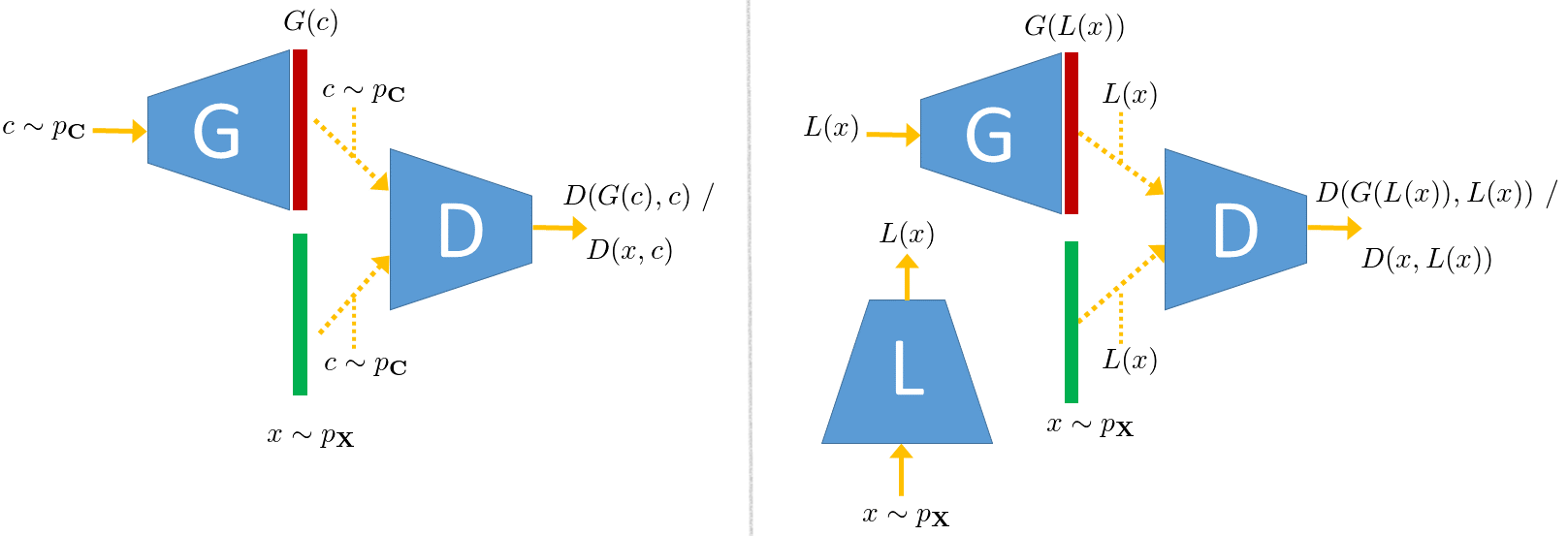}
  \vspace{-12pt}
  \caption{Conventional cGAN formulation on the left forms the supervised cGAN objective. On the right is the proposed unsupervised GAN objective with the labeller network L.}
  \label{fig:method_schematic}
\vspace{-12pt}
\end{figure*}

As discussed above, standard training of cGANs requires labelled training data to provide pairs of data sample and corresponding conditional (label) for the discriminator input. For a semi-supervised technique we need a way to incorporate unsupervised examples during training. We do this by introducing an additional \textit{labeller} network, which generates conditional labels from unsupervised inputs. For a class conditional cGAN, this labeller network could be a classifier network (e.g. Resnet \cite{resnet_he_2016}), while for a cGAN conditioned on semantic maps, this could be a segmentation network (e.g. DeepLabv3+ \cite{deeplabv3plus2018}). This labeller network is trained jointly with the generator and discriminator. See Figure~\ref{fig:method_schematic} for a schematic of our method.

We now describe the optimization objective of \sscgan{}. Let $\Sset = \{(\xb, \cb)\}$ be the set of supervised labelled data containing pairs of data samples and their labels. Let $\Uset = \{\xb\}$ be the set of unsupervised data points without any labels. The various terms for the optimization objective are:

\medskip\noindent\textbf{Supervised GAN objective:} This is the standard cGAN term $V_c(D,G)$ of Equation~\ref{eq:cgan_objective} and acts on the supervised dataset \Sset. The discriminator D is trained to maximize, while the generator G is trained to minimize this objective.

\medskip\noindent\textbf{Supervised Labeller loss:} This loss term uses the supervised set $\Sset$ to makes sure that the labeller network L is steered towards predicting accurate labels for the data samples:
\begin{align}
\label{eq:labeller_sup_loss}
V_{labeller}(L) = \E_{(\xb,\cb)\sim \pxc} l(L(\xb), \cb),
\end{align}
where $l(L(\xb), \cb)$ denotes the loss for individual samples. For example, in the case of a class conditional cGAN, the labeller network L is a classifier network and $l()$ may correspond to the cross-entropy loss. Similarly, for a cGAN conditioned on semantic maps, L is a segmentation network and $l()$ may correspond to a pixel-wise loss. Note that our framework is independent of the choice of $l()$.

\medskip\noindent\textbf{Unsupervised GAN objective:} This is the objective which incorporates the unsupervised dataset \Uset, and is the main contribution of this work. We construct this objective to be similar to $V_c(D,G)$, by leveraging the labeller network $L$ to generate conditional labels for the unsupervised samples. One candidate objective for this unsupervised data would be:

{\small
\begin{multline}
\label{eq:cgan_objective_unsup_conditional_sampling}
\tilde{V}_c^u(D, G, L) := \\
\E_{\xb\sim \px}[log D(\xb, L(\xb))] + \E_{\cb\sim \pc}[log(1 - D(G(\cb), \cb))].
\end{multline}
}%
However, note that to compute the second term of $\tilde{V}_c^u$, we need to sample from the true conditional distribution $\pc$. This may be possible sometimes, e.g. in the case of class conditionals where we may know the underlying distribution of the various classes (say 15\% cats, 35\% cars, etc.). However in most cases, where we have access to only unlabelled data, it is not possible to access the distribution of these labels (conditionals). For example, in the case of semantic map conditionals, we can not sample the semantic maps for estimating equation~\ref{eq:cgan_objective_unsup_conditional_sampling} unless we have access to the labelled semantic maps. To solve this, we propose the following objective:
\begin{multline}
\label{eq:cgan_objective_unsup_full}
V_c^u(D, G, L) := \E_{\xb\sim \px}[log D(\xb, L(\xb))] + \\
\E_{\xb\sim \px}[log(1 - D(G(L(\xb)), L(\xb)))].
\end{multline}
Here, the labeller network $L$ is also used to generate conditional inputs needed by the generator (see Figure~\ref{fig:method_schematic}).

Again, similar to the supervised cGAN objective, the discriminator D is trained to maximize this unsupervised objective, while the generator G is trained to minimize this. However the important question here is: \textit{what should the labeller network L do?} Unfortunately, the traditional GAN formulation does not provide a good intuition to answer this, but Wasserstein GAN formulation comes to the rescue here. In the Wasserstein GAN formulation, D acts as a critic and is trained to maximize an objective similar to traditional GAN:
\begin{align}
\label{eq:wgan_critic_objective}
W := \E_{\xb\sim \px}[D(\xb)] - \E_{\xb\sim \pg}[D(\xb)].
\end{align}
It turns out that under certain conditions (see \cite{arjovsky2017wasserstein}), once maximized, the objective W approximates the Wasserstein distance between the two distributions $\px$ and $\pg$. Since the purpose of the generator is to make the generated distribution $\pg$ close to the real distribution $\px$, it should minimize this distance W. If we look at our objective $V_c^u(D, G, L)$ with ``\textit{Wasserstein glasses}'', once maximized for the discriminator, it corresponds to the Wasserstein distance between the two joint distributions corresponding to $(\xb, L(\xb))$ and $(G(L(\xb)), L(\xb))$. Since we want the two distribution to converge, it is clear that we should minimize this objective w.r.t $L$ as well. Note that although we used the Wasserstein formulation to motivate the above discussion, our result holds for even the standard GAN formulation. In fact, our labeller's formulation is very similar to that of the encoder in the ALI and BiGAN papers~\cite{dumoulin2016ali,donahue2016bigan}, where the GAN objective is also minimized w.r.t the encoder. Please see these papers for detailed proofs, which apply to our formulation as well.

\medskip\noindent\textbf{Final Objective:} Putting it all together, the final objective is obtained by combining $V_c(D,G)$, $V_{labeller}(L)$ and $V_c^u(D, G, L)$:
\begin{multline}
\label{eq:full_objective}
V_{full}(D, G, L) := \lambda_{1} V_c(D,G) + \\
\lambda_{2} V_{labeller}(L) + \lambda_{3} V_c^u(D, G, L),
\end{multline}
where $\lambda_1, \lambda_2, \lambda_3$ are hyperparameters. The three networks D, G, and L are optimized in a minimax fashion, i.e:
\begin{align}
\label{eq:ss_gan_minimax}
\min_{G,L} \max_{D} V_{full}(D, G, L).
\end{align}

\medskip\noindent\textbf{Gumbel Softmax:} Note that the output of the labeller network $L$ is typically discrete (e.g. semantic labels for each pixel) implemented via an argmax layer. This poses a problem for the joint training of $L$ along with rest of the GAN as the argmax layer is non differentiable. To solve this, we replaced the argmax layer with a Gumbel Softmax layer \cite{gumbel2017}, which allows us to sample discrete labels, while at the same time also allows for estimation of gradients for the backward pass.

\subsection{Unsupervised Regularizer}
In this section we analyze the role of the unsupervised objective of Equation~\ref{eq:cgan_objective_unsup_full} in training of \sscgan{}. Let, $\Ls$ and $\Gs$ be the optimal L and G at the end of optimizing Equation~\ref{eq:ss_gan_minimax}. As discussed above, at end of the minimax optimization of GAN objective, the two distributions being compared by the discriminator can be assumed to converge, i.e. we can assume:
\begin{align}
\label{eq:unsupervised_distribution_match}
\px(\xb)\pLs(\cb|\xb) = \pgs(\xb|\cb)\pLs(\cb),
\end{align}
where $\px(\xb)\pLs(\cb|\xb)$ is the probability of sampling the pair $(\xb, L(\xb))$, while $\pgs(\xb|\cb)\pLs(\cb)$ is the probability of sampling the pair $(G(L(\xb)), L(\xb))$ in Equation~\ref{eq:cgan_objective_unsup_full}. Here $\pLs(c) = \int_{\Sset_{x} \cup \Uset} \pLs(\cb|\xb)p(\xb)d\xb$ is the distribution on conditionals induced by the labeller  when sampling over the reals. Here $\Sset_x = \{\xb | (\xb, \cb) \in \Sset\}$, and we also similarly define $\Sset_c = \{\cb | (\xb, \cb) \in \Sset\}$ for later reference.

Now, consider the points in the supervised domain, i.e. $(\xb, \cb) \in \Sset$. If we make the (reasonable) assumption that for points in the supervised domain, the labeller is able to predict the true labels accurately (by virtue of loss in Equation~\ref{eq:labeller_sup_loss}), and the generator is able to generate samples satisfying the input conditionals accurately (by virtue of objective in Equation~\ref{eq:cgan_objective}), we can say
\begin{align}
\pLs(\cb|\xb) &= \pxc(\cb|\xb) \label{eq:perfect_labeller} \\
\pgs(\xb|\cb) &= \pxc(\xb|\cb) \label{eq:perfect_g} ,
\end{align}
where $\pxc$ denotes the true distributions. Substituting \ref{eq:perfect_labeller},~\ref{eq:perfect_g} in Equation~\ref{eq:unsupervised_distribution_match}:
\begin{align}
\label{eq:unsupervised_distribution_match_1}
\px(\xb)\pxc(\cb|\xb) &= \pxc(\xb|\cb)\pLs(\cb) \\
\implies \pLs(\cb) &= \frac{\px(\xb)\pxc(\cb|\xb)}{\pxc(\xb|\cb)} \\
\implies \pLs(\cb) &= \frac{\pxc(\cb,\xb)}{\pxc(\xb|\cb)} \\
\implies \pLs(\cb) &= \pc(\cb)
\end{align}
Using the definition of $\pLs(\cb)$, we get
\begin{align}
\label{eq:unsupervised_distribution_match_2}
\int_{\Sset_{x} \cup \Uset} \pLs(\cb|\xb)p(\xb)d\xb = \pc(\cb)
\end{align}
That is, the optimal labeller $\Ls$ is such that, for points in the supervised set ($\cb \in \Sset_c$), when $\Ls$ is marginalized over the full domain ($\Sset_{x} \cup \Uset$), it gives the true probability of the conditionals for these supervised points. Thus the supervised points constraint the labeller not only in the supervised region $\Sset_{x}$, but also in the unsupervised region $\Uset$, such that the marginalized probability over the full region equals the true probability at these supervised points. In some sense, this acts a regularizer on the labeller network, which in turn helps the cGAN train better.

\subsection{Two-pass Inference}
\label{sec:two-pass-inference}

The generator G of our \sscgan{} takes two different sets of conditional inputs during training. For the supervised set $\Sset$, it take the real conditionals $\Sset_c$ (see eq~\ref{eq:cgan_objective}), while for the unsupervised set $\Uset$ it takes the conditionals inferred by the labeller L from the reals, i.e. $\Uset_c = \{L(\xb) | \xb \in \Uset\}$ (see eq~\ref{eq:cgan_objective_unsup_full}). Since we train our models with sparse labelled data, i.e. $|\Sset| \ll |\Uset|$, the generator G tends to perform better for input conditionals drawn from the distribution of $\Uset_c$ as compared to that of $\Sset_c$.

To incorporate this observation during our inference procedure, we follow a two-pass scheme. In the first pass, we generate a fake sample from the input conditional as usual, i.e. $\xb_{fake}^1 = G(\cb_{input})$, where $\cb_{input}$ is the conditional input. Next, we pass this fake output through the labeller L to generate a synthetic conditional $L(\xb_{fake}^1)$ which is closer to the distribution of $\Uset_c$, as compared to the original input $\cb_{input}$. This synthetic conditional is then passed to the generator for the final fake output $\xb_{fake}^2 = G(L(\xb_{fake}^1))$. We found significant improvement in the quality of generated samples with this two-pass scheme. Please see Figure~\ref{fig:camhq_onepass_vs_twopass} in Appendix for detailed comparison.
\section{Experiments}
\label{sec:experiments}

We implemented our \sscgan{} framework on two different cGAN tasks -- semantic image synthesis and class conditional image synthesis. We discuss only the semantic images synthesis task in the main paper. Please refer to the supplementary material for the experiments on class conditional synthesis. The source code of our implementations will be released soon.

\subsection{Semantic Image Synthesis}
\label{sec:semantic_synthesis_expts}
Semantic image synthesis looks at the specific image-to-image translation task of synthesizing photo-realistic images from semantic maps. For our evaluation, we incorporate the \sscgan{} framework into the recently proposed SPADE network \cite{park2019spade}. For the labeller network L, we use DeepLabv3+ segmentation network \cite{deeplabv3plus2018}. Our integration required only $\approx250$ lines of code.
We used the following two datasets for evaluation:
\begin{itemize}
    \item \textit{\camhq{}} \cite{lee2019maskgan} is a dataset of annotated celebrity faces built on the CelebA-HQ \cite{karras2017progressive} dataset. The authors added precise pixel-wise hand annotations for facial components, such as nose, eyes, etc. The dataset contains 30,000 labelled images. We split the dataset into 29,000 training and 1000 test sets.
    
    \item \textit{CityScapes} \cite{cordts2016cityscapes} is a dataset of 3500 street images of German cities, containing dense pixel-wise annotations such as road, building, car, etc. The dataset is split into 3000 training and 500 test sets.
\end{itemize}
For each dataset, we use only a subset of the training set for forming our supervised set \Sset{} with labels, while from the remaining we take only the images without labels to form our unsupervised set \Uset{}.

\medskip\noindent\textbf{Baselines:} Our first baseline is the \textit{Fully supervised} baseline, i.e. vanilla SPADE, where we train the SPADE network with the full supervised training set. This baseline can be expected to give the best results as it uses the entire supervised training data. The second baseline is what we call the \textit{\bltwocap{}} baseline, where we first train the labeller network with the supervised subset \Sset{}, and use that to generate labels for all images in \Uset{}. The SPADE network is then trained as usual with these synthetic labels as well as those of \Sset{}.

\begin{figure*}[h]
  \centering
  \vspace{-28pt}
  \begin{tabular}{ccc}
    \includegraphics[width=.28\textwidth]{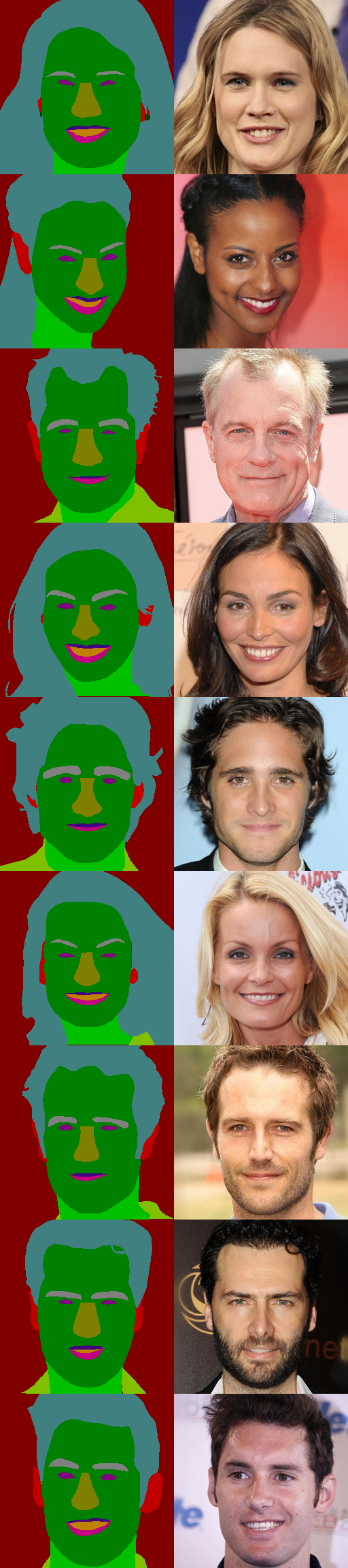} &
    \includegraphics[width=.14\textwidth]{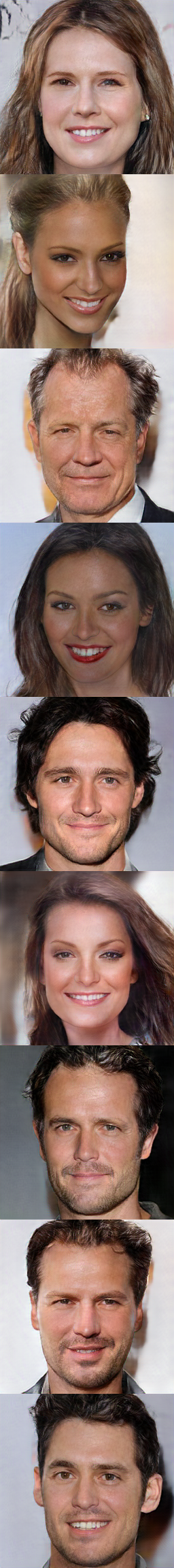}  &
    \includegraphics[width=.28\textwidth]{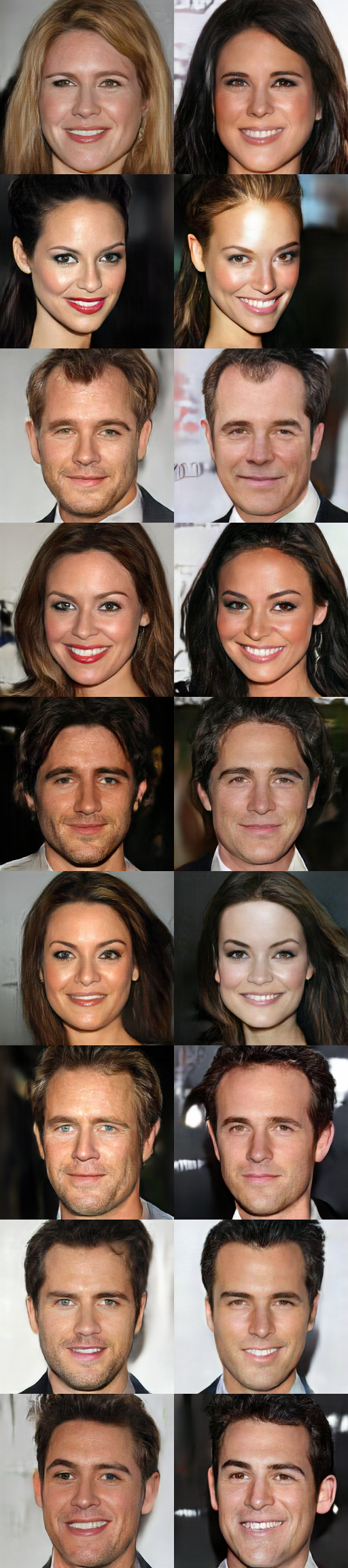} \\
    {\scriptsize Label, Ground Truth} 
    & {\scriptsize Fully Supervised} 
    & {\scriptsize \sscgan{}: 5, 25 labelled pairs}
  \end{tabular}
  \vspace{-8pt}
  \caption{Synthesis results (cherry-picked) for \camhq{} dataset on test set semantic maps. Shown are results from fully-supervised SPADE and \sscgan{} with 5 and 25 labelled pairs. \sscgan{} produces realistic results, while accurately respecting the semantic layout (even when trained with just 5 supervised pairs!).}
  \label{fig:camhq_fully_sup_vs_ours}
\end{figure*}

\begin{figure*}[h]
  \centering
  \vspace{-28pt}
  \begin{tabular}{ccc}
    \includegraphics[width=.28\textwidth]{figures/camhq/grid_reals.png} &  
    \includegraphics[width=.28\textwidth]{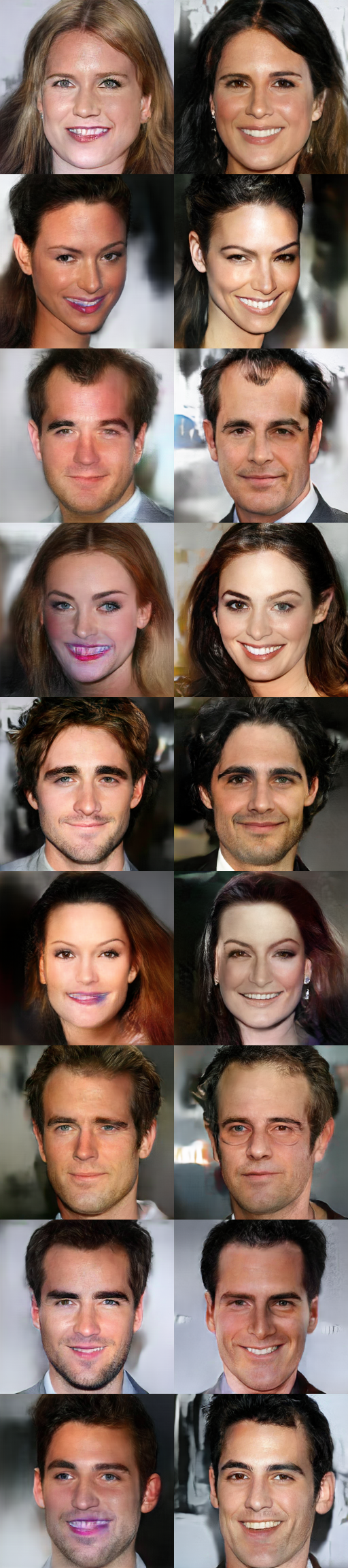}   &
    \includegraphics[width=.28\textwidth]{figures/camhq/grid_ours.png} \\
    {\scriptsize Label, Ground Truth} &
    {\scriptsize \bltwocap{}: 5, 25 labelled pairs} & {\scriptsize \sscgan{}: 5, 25 labelled pairs}
  \end{tabular}
  \vspace{-2pt}
  \caption{Comparison of \bltwocap{} to \sscgan{} synthesis results for \camhq{} test set on the same test inputs as Figure~\ref{fig:camhq_fully_sup_vs_ours}. Results are shown for training with 5 and 25 labelled pairs. \bltwocap{} struggles with synthesis quality, while \sscgan{} synthesizes realistic images. Interestingly, both methods respect the semantic layout well.}
  \label{fig:camhq_bl2_vs_ours}
\end{figure*}

\begin{figure*}[h]
  \centering
  \begin{tabular}{c@{\hspace{.7\tabcolsep}}c@{\hspace{.7\tabcolsep}}c@{\hspace{.7\tabcolsep}}c}
    \includegraphics[width=.271\textwidth]{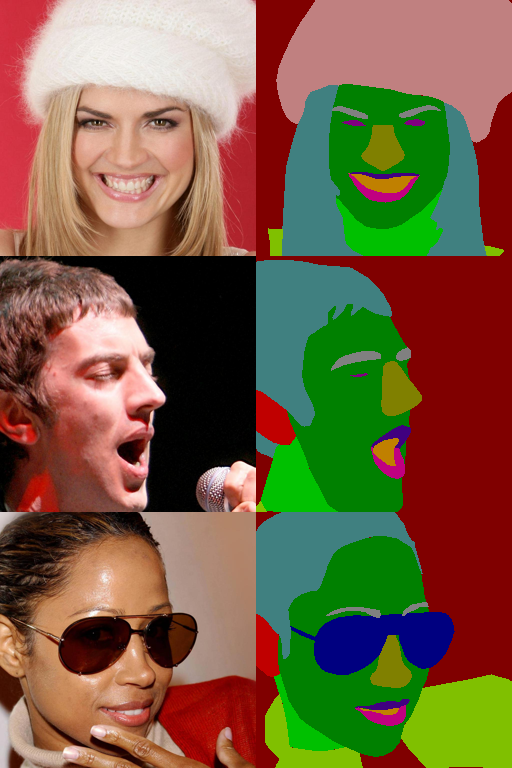} &   \includegraphics[width=.136\textwidth]{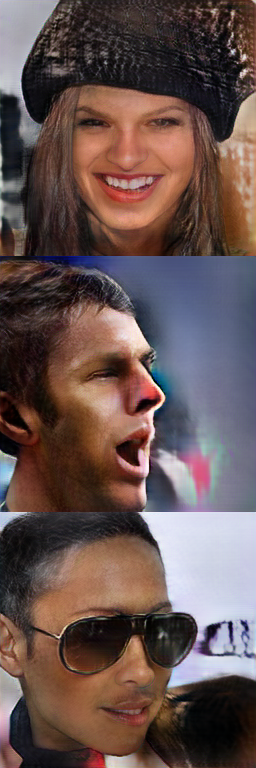}  &
    \includegraphics[width=.271\textwidth]{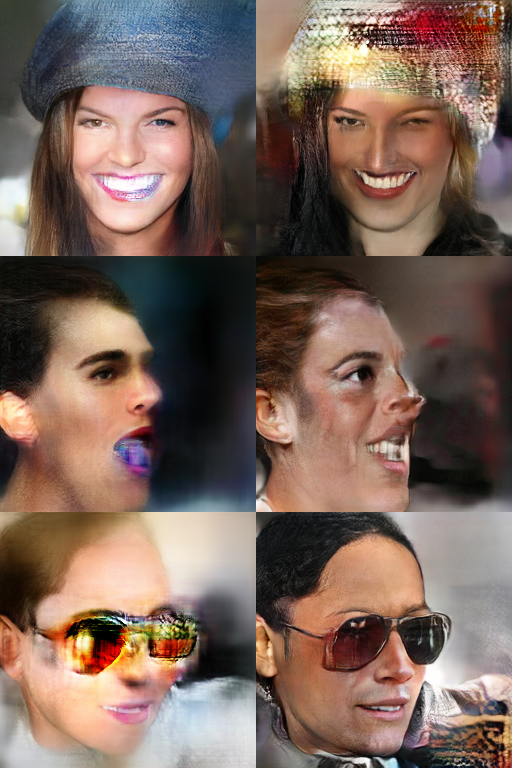}  &
    \includegraphics[width=.271\textwidth]{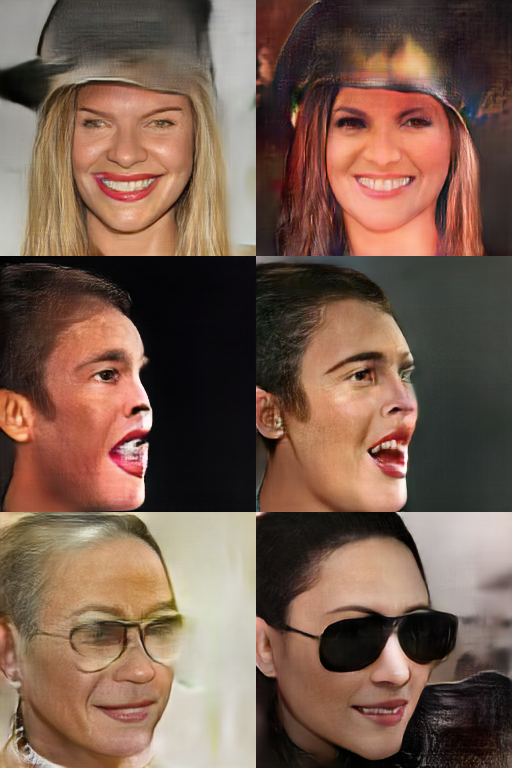} \\
    {\scriptsize Label, Ground Truth} & {\scriptsize Fully} & {\scriptsize \bltwocap{}:} & {\scriptsize \sscgan{}:} \\
    & {\scriptsize Supervised} & {\scriptsize 5, 25 labelled pairs} & {\scriptsize 5, 25 labelled pairs} \\
  \end{tabular}
  \vspace{-2pt}
  \caption{Failure cases on \camhq{} dataset. Results are shown with 5, 25 labelled pairs. All methods perform badly when the input semantic map is not well represented in the training distribution. The fraction of training images with cap, glasses and profile view are quite less, and as a result the networks struggle with generating realistic results even for the fully supervised case.}
  \label{fig:camhq_failure_cases}
\end{figure*}

\medskip\noindent\textbf{Synthesis Results:}
For the \camhq{} dataset we run two sets of experiments. In the first one we use only 5 labelled pairs out of the 29000 training set as the supervised set, and use only the unpaired images from the remaining 28995 pairs. These 5 images (See figure~\ref{fig:showcase_5_paired_results}) were hand picked, so that we cover a few different face angles, as well as persons wearing glasses, cap, etc. In the second experiment, we train using 25 labelled pairs (chosen randomly) and 28975 unpaired images. Both experiments were trained at a resolution of 256x256. Figure~\ref{fig:camhq_fully_sup_vs_ours} shows cherry-picked results from our method for the three experiments compared to the fully-supervised (vanilla SPADE) baseline. Comparison with the \bltwo{} is shown in figure~\ref{fig:camhq_bl2_vs_ours}. All results are with semantic map inputs from the test dataset which is not used during training. It is interesting to note that even with a sparse labelled set of just 5 images, \sscgan{} performs quite well, especially for semantic map inputs which are qualitatively covered in the 5 training pairs. However, for semantic maps not covered in the training distribution (e.g. none of the 5 training pairs have teeth), the results may exhibit artifacts. See figure~\ref{fig:camhq_failure_cases} for examples of such failure cases. More results with randomly selected samples from the test set are shown in the supplementary material.

Note that we found that for this dataset, the Frechet Inception Distance (FID) scores \cite{heusel2017ganstturfid} did not correlate with the visual quality of the synthesized results\footnote{This could be because the inception network is trained on a dataset not representative of \camhq{} dataset.}. For example, in Figure~\ref{fig:camhq_bl2_vs_ours} the 5-paired \bltwo{} (1st column) which has very bad visual quality with a lot of artifacts, gave superior FID score compared to 5-paired \sscgan{} (4th column) which is of much superior visual quality. Also see Figure~\ref{fig:fid_camhq_5_bl2_vs_ours} in supplementary. Thus we do not report FID scores for this dataset.

\begin{figure*}[h]
  \centering
  \begin{tabular}{c@{\hspace{0pt}}c @{\hspace{.92\tabcolsep}} c@{\hspace{0pt}}c @{\hspace{.92\tabcolsep}} c@{\hspace{0pt}}c}
    \includegraphics[width=.159\textwidth]{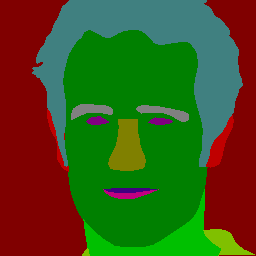} &
    \includegraphics[width=.159\textwidth]{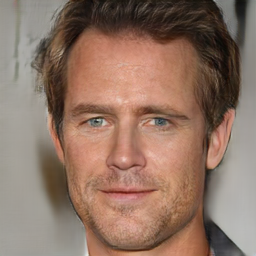} &
    
    \includegraphics[width=.159\textwidth]{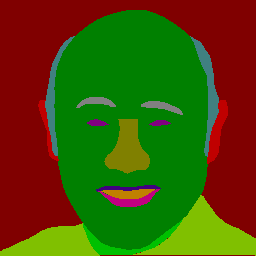} &
    \includegraphics[width=.159\textwidth]{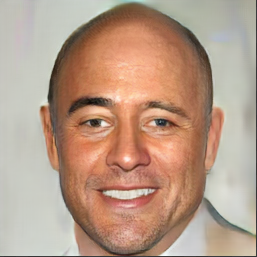} &
    
    \includegraphics[width=.159\textwidth]{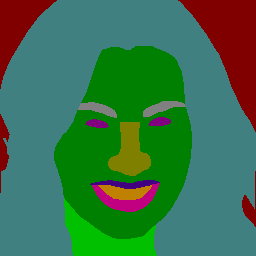} &
    \includegraphics[width=.159\textwidth]{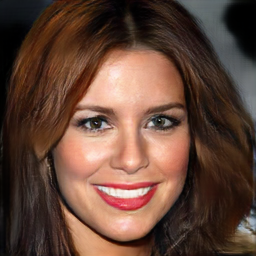}
    
    \\
    \includegraphics[width=.159\textwidth]{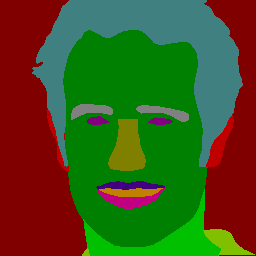} &
    \includegraphics[width=.159\textwidth]{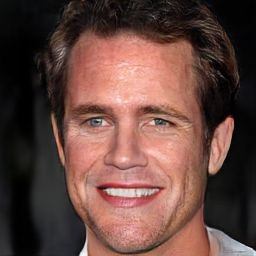} &
    
    \includegraphics[width=.159\textwidth]{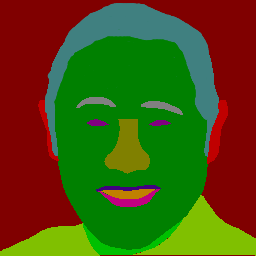} &
    \includegraphics[width=.159\textwidth]{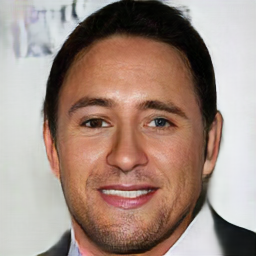} &
    
    \includegraphics[width=.159\textwidth]{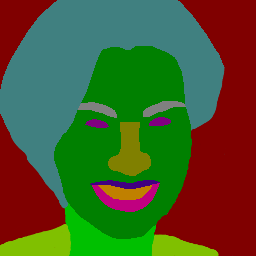} &
    \includegraphics[width=.159\textwidth]{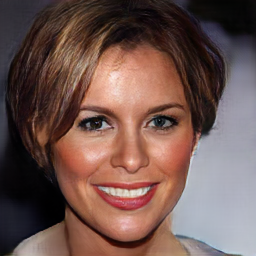}
    
    \\
    
    \multicolumn{2}{c}{\scriptsize Edited Lips} & \multicolumn{2}{c}{\scriptsize Edited Hair} & \multicolumn{2}{c}{\scriptsize Edited Hair}
    
    \\
    
  \end{tabular}
  \caption{Custom editing. We took semantic maps from the test set of \camhq{} and edited the map with a paint tool. Top row shows the original semantic map with the corresponding synthesized results. Bottom row shows the edited semantic map and synthesized results. We used \sscgan{} trained with only 5-pairs for generating the synthesized results. As shown, even with 5 supervised pairs, the network is able to generate realistic results while accurately respecting the semantic layout.
  }
  \label{fig:custom_edits}
\end{figure*} 

\begin{figure*}[h]
  \centering
  \resizebox{\textwidth}{!}{
  \begin{tabular}{cccc}
    \includegraphics[width=.4\textwidth]{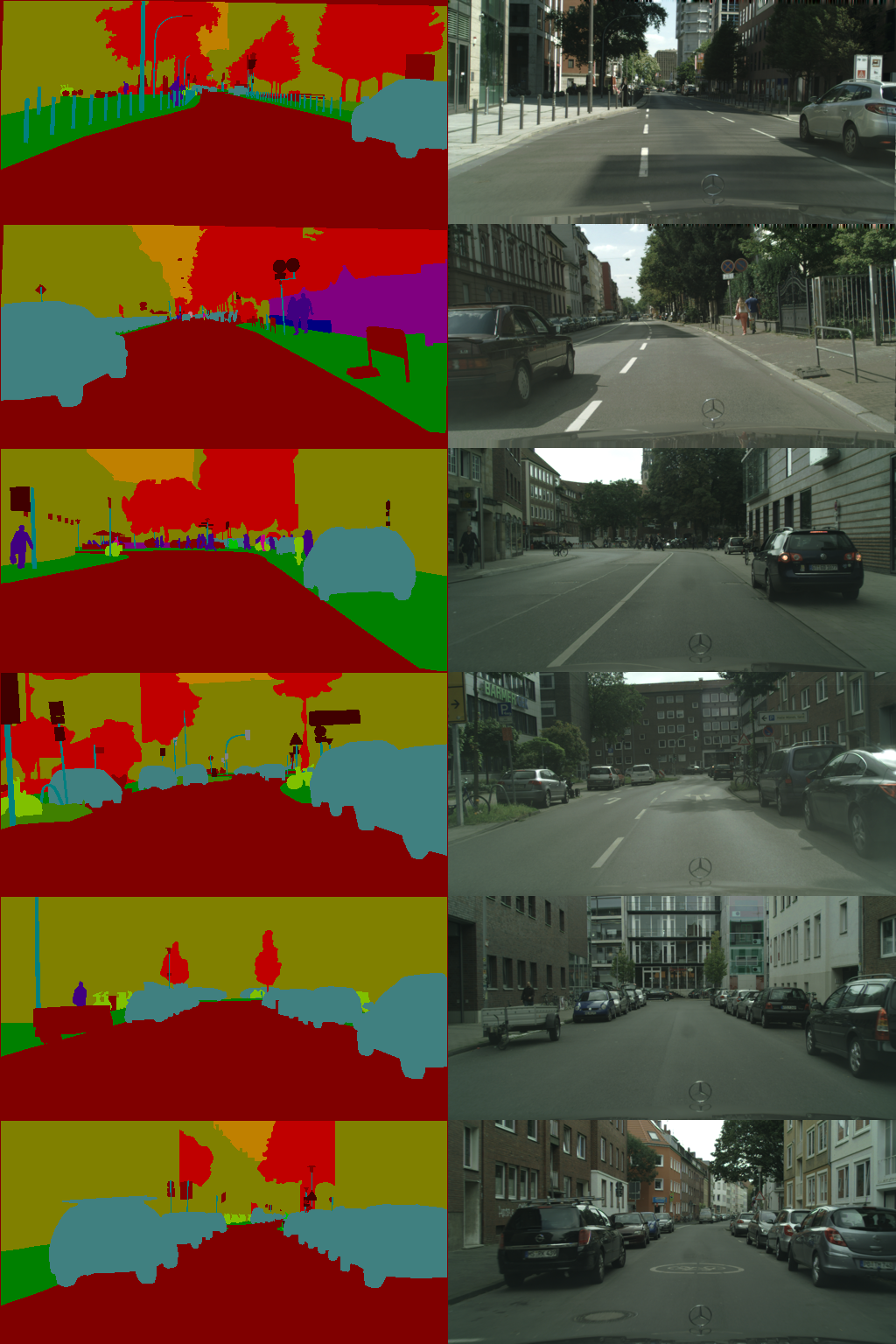} &   \includegraphics[width=.2\textwidth]{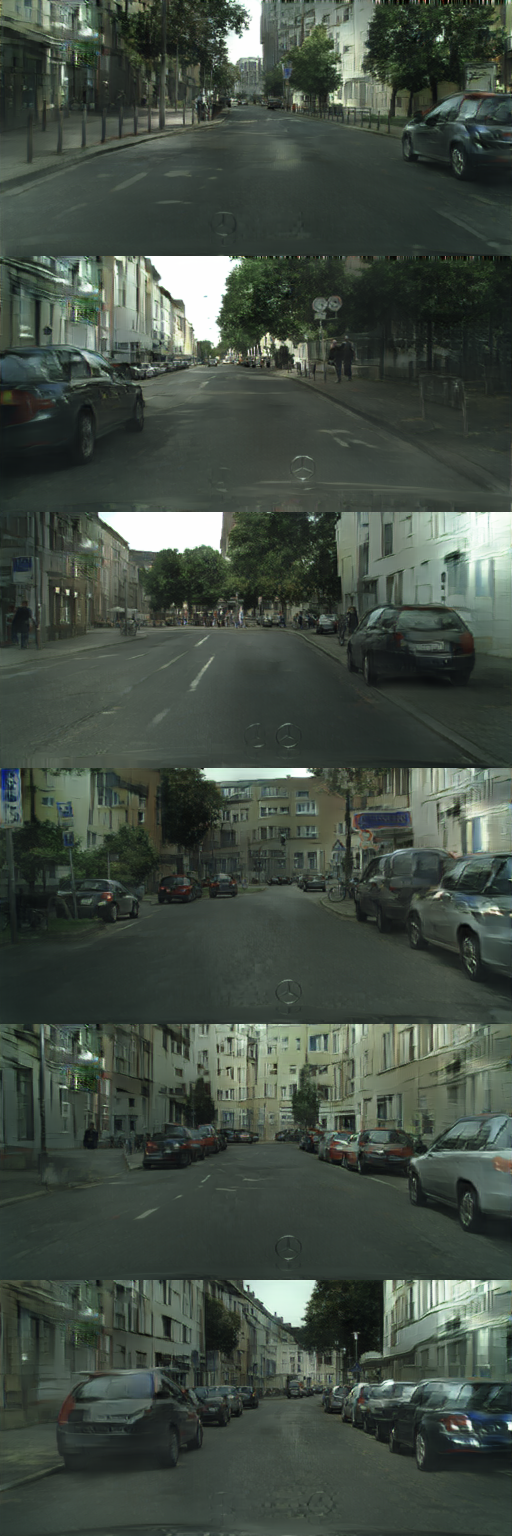}   &
    \includegraphics[width=.4\textwidth]{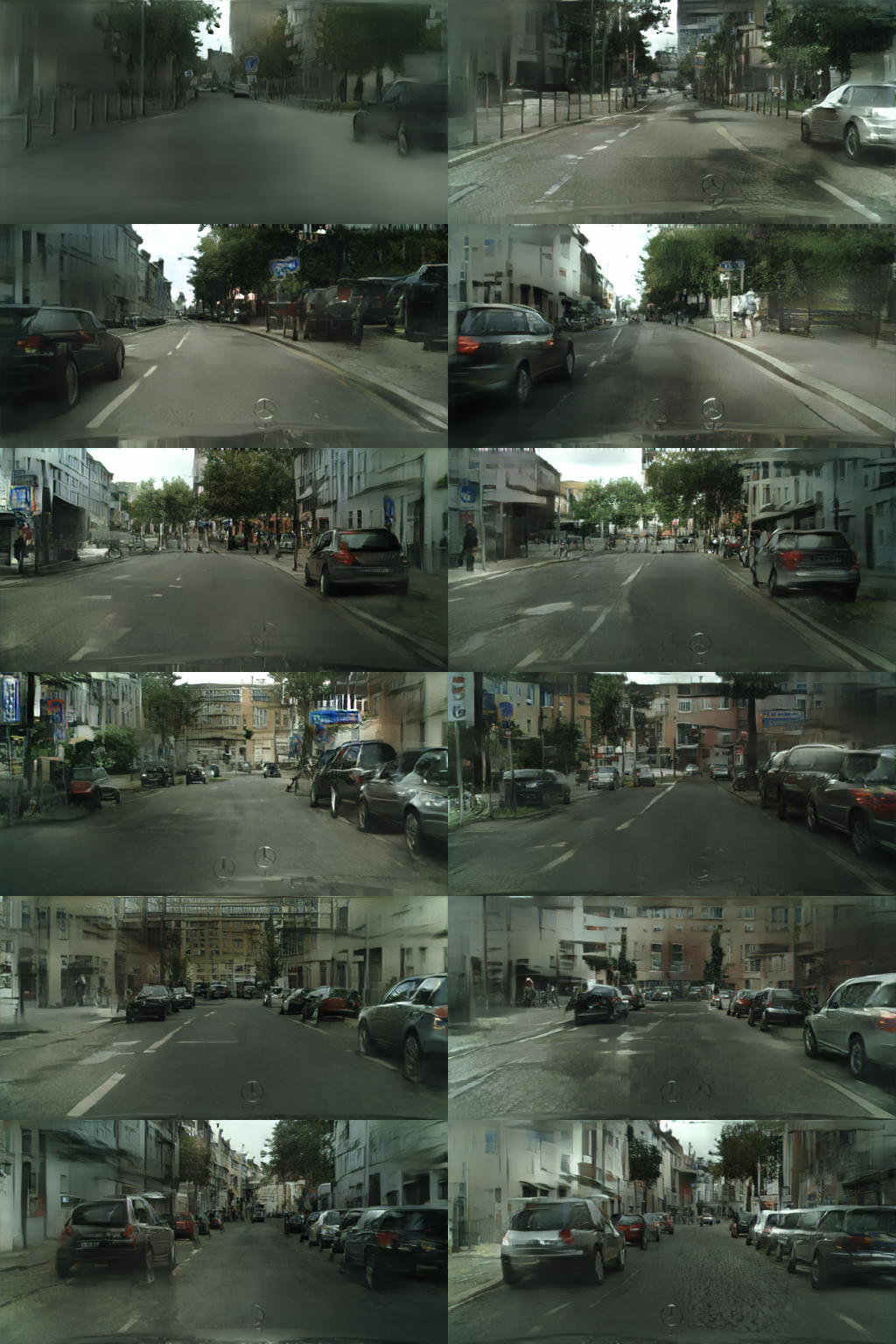} &
    \includegraphics[width=.4\textwidth]{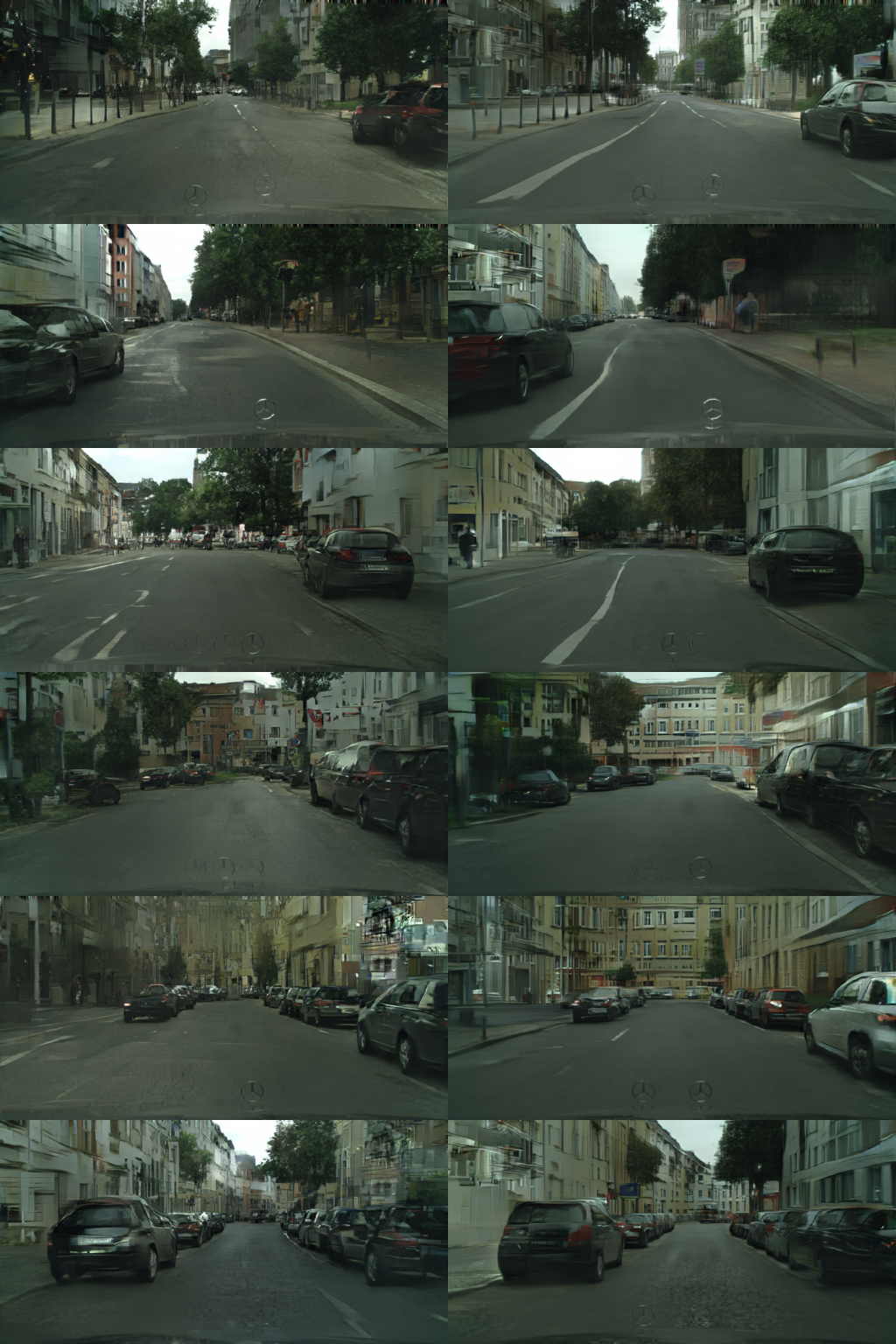} \\
    {\scriptsize Labels, Ground Truth} & {\scriptsize Fully Supervised} & {\scriptsize \bltwocap{}: 100, 500 labelled pairs} & {\scriptsize \sscgan{}: 100, 500 labelled pairs} \\
    & {\scriptsize FID: 58.76} & {\scriptsize FID: 73.71, 69.13} & {\scriptsize FID: 65.06, 53.83}
  \end{tabular}
  }
  \caption{Synthesis results (cherry-picked) for CityScapes dataset on semantic maps from the test set. Results are shown for fully-supervised SPADE, \bltwo{} and \sscgan{} along with the FID scores computed on the test set.}
  \label{fig:city_results}
\end{figure*}

For the CityScapes dataset we run two experiments with 100 and 500 labelled pairs (randomly chosen) from the 3000 training set as the supervised set. The remaining images (2900 and 2500, respectively) form the unsupervised set. All experiments were trained at a resolution of 512x256. Figure~\ref{fig:city_results} show cherry-picked results comparing our method to the baselines. All results are with semantic maps from the test dataset which is not used for training. Also shown are the FID scores computed on the test set. More results with randomly selected samples from the test dataset are shown in the supplementary material.

\medskip\noindent\textbf{Human Perceptual Study:} To evaluate the visual quality of our synthesis, we carried out a perceptual study, similar to \cite{isola2017pix2pix}, on Amazon Mechanical Turk (AMT). The study displayed an image (real or synthesized) to a turker for one second, after which the turker was given indefinite time to label it as real or fake. Each turker was shown 42 images (7 per experiment), all chosen at random. Synthesized images were generated using the test set labels only. We report the \textit{real perception rate} (RPR) for each experiment, defined as the percentage of images reported as real to the number of images labelled. To remove insincere turkers, we discard entries of any turker who labelled any real image as fake. %
The RPR scores are reported in Table \ref{tab:amt_scores}.

\begin{table}[]
\centering
\resizebox{0.475\textwidth}{!}{%
\begin{tabular}{ccc}

\toprule
\textbf{Dataset}  & \textbf{Model} & \textbf{RPR} \\
\midrule
\multirow{5}{*}{\camhq{}} & \bltwocapshort{} (5 pairs)   & 32.5 \\ 
                          & \bltwocapshort{} (25 pairs)  & 43.8 \\ \cline{2-3}
                          & \sscgan{} (5 pairs)          & 49.6 \\ 
                          & \sscgan{} (25 pairs)         & 66.0 \\ \cline{2-3}
                          & Fully Supervised             & 70.0 \\
\midrule

\multirow{5}{*}{Cityscapes} & \bltwocapshort{} (100 pairs)  & 54.2 \\ 
                            & \bltwocapshort{} (500 pairs)  & 54.3 \\ \cline{2-3}
                            & \sscgan{} (100 pairs)         & 65.2 \\ 
                            & \sscgan{} (500 pairs)         & 66.3 \\ \cline{2-3}
                            & Fully Supervised              & 73.6 \\
\bottomrule

\end{tabular}
}
\caption{RPR \textit{(real perception rate)} scores from AMT study on the test splits. %
As shown, \sscgan{} significantly outperforms the \bltwocap{} for both datasets. For the \camhq{} dataset, even with just 25 labeled pairs, \sscgan{} RPR is very close the the fully supervised case which is trained with 29000 pairs. }
\label{tab:amt_scores}
\end{table}

\medskip\noindent\textbf{Conditional Mapping Metrics:} We also measure how accurately \sscgan{} respects the semantic map inputs. We report the standard semantic segmentation scores (mean IoU and overall pixel accuracy) in Tables~\ref{tab:semantic_segmentation_scores_camhq},~\ref{tab:semantic_segmentation_scores_city}. The scores were computed by first training a high accuracy segmentation network (we used DeepLabv3+ \cite{deeplabv3plus2018}) on the full supervised set, and then computing the segmentation maps of the generated images. As can be seen in the tables, \sscgan{} performs quite well. The performance of \camhq{} 5 labelled pair network is especially impressive, given the small number of supervised data it sees.

\begin{table}[h]
\centering
\resizebox{0.425\textwidth}{!}{

  {\setlength{\extrarowheight}{1pt}
  \begin{tabular}{ccc}
  \toprule
  Experiment & mIoU & Pixel \\
             &      & Accuracy \\
  \midrule
  \bltwocap{} (5 pairs)  & 65.72 & 92.84 \\
  \bltwocap{} (25 pairs) & 63.43 & 91.94 \\
  \midrule
  \sscgan{} (5 pairs)    & 69.95 & 93.78 \\
  \sscgan{} (25 pairs)    & 72.70 & 94.92 \\
  \midrule
  Fully-supervised       & 74.88 & 95.08 \\
  \bottomrule
  \end{tabular}
  }
  }
  \caption{Semantic segmentation scores on the test split for \camhq{} dataset. \sscgan{} outperforms baseline and comes close to the fully-supervised run even with few labelled pairs. Interestingly \bltwocap{} with 25 pairs under-performs compared to the 5 pairs run, most likely due to the carefully hand picked 5 pairs compared to the randomly chosen 25 pairs.}
  \label{tab:semantic_segmentation_scores_camhq}
\end{table}

\begin{table}[h!]
\small
\centering
  \resizebox{0.425\textwidth}{!}{
  {\setlength{\extrarowheight}{1pt}
  \begin{tabular}{ccc}
  \toprule
  Experiment & mIoU & Pixel \\
             &      & Accuracy \\
  \midrule
  \bltwocap{} (100 pairs) & 41.84 & 82.50 \\
  \bltwocap{} (500 pairs) & 44.96 & 83.70 \\
  \midrule
  \sscgan{} (100 pairs) & 44.23 & 84.92 \\ 
  \sscgan{} (500 pairs) & 49.40 & 85.80 \\
  \midrule
  Fully-supervised  & 51.52 & 87.29 \\
  \bottomrule
  \end{tabular}}
}
  \caption{Semantic segmentation scores on the test split for CityScapes dataset. \sscgan{} outperforms baseline and comes close to the fully-supervised run even with the small number of labelled pairs.}
  \label{tab:semantic_segmentation_scores_city}
\end{table}

\subsection{Custom Editing}
To demonstrate robustness of the learnt network, we took a few semantic maps from our test set and made custom edits to them with a paint tool. The results are shown in Figure~\ref{fig:custom_edits}. As shown, the network robustly follows the change in semantic maps and produces realistic output.

\section{Related Work}
\medskip\noindent\textbf{GAN:} Since their introduction \cite{goodfellow2014generative}, GANs have been remarkably successful in generating complex real world data distributions. Apart from image generation \cite{karras2017progressive,karras2019styleprogressive2,radford2015dcgan,miyato2018spectral,zhang2018sagan}, which is the primary concern in this work, GANs have also enabled other applications such as representation learning \cite{chen2016infogan,radford2015dcgan}, image manipulation \cite{zhu2016manipulationmanifold}, etc.

\medskip\noindent\textbf{cGAN:} Conditional GANs \cite{mirza2014conditionalgan,miyato2018cgansprojection} provide control over the synthesis process by conditioning the output on input conditions. cGANs have been successfully used in multiple scenarios such as class conditional generation \cite{brock2018biggan,kavalerov2019mhingegan}, image-to-image translation \cite{isola2017pix2pix,wang2018pix2pixhd,papadopoulos2019makepizza,park2019spade,lee2019maskgan}, super resolution \cite{ledig2017superresolution}, colorization \cite{iizuka2016lettherebecolor,zhang2016colorful,zhang2017colorizationdeeppriors}, image completion \cite{pathak2016contextencoder,iizuka2017imagecompletion}, etc.

\medskip\noindent\textbf{Unpaired Image to Image translation:} Although cGANs provide control over the synthesis process, they come at the cost of requiring large amounts of labelled data. To avoid this cost, several techniques have been tried. \cite{zhu2017cyclegan} targets the problem of translating images from one domain to another in an unsupervised fashion by introducing a cycle-consistency loss between forward and inverse generators. \cite{liu2019funit} solves a similar problem of unsupervised image translation from one class to another, but with the additional constraint of using only a few images of the target class.%

\vspace{4mm}

\medskip\noindent\textbf{Semi-Supervised training of cGANs:} Although unpaired image-to-image translation methods don't require any labelled data, they provide only high level control such as domain translation. Our work lies in the category of semi-supervised training of conditional GANs, which provide fine control of the synthesis process similar to standard cGANs, but require lot less data. This line of work has seen recent interest. S3GAN \citep{lucic2019highfidfewerlabels} achieves state of the art results for ImageNet using only 10\% of the labels. The paper employs a mix of self-supervised (e.g. predicting rotation) and semi-supervised techniques to generated labels for all datapoints, which are then used to train the cGAN. Their method, however is specific to class conditioned cGANs as it is based on the property of a GAN discriminator being similar to a classifier. Our method, on the other hand, is generic and works for class conditional and semantic image synthesis alike. \cite{stoller2019incompletefactorized} uses a smart factorization technique, and uses multiple discriminators to estimate the different factors. Their factorization allows partial data (e.g. missing labels) to be utilized. However, one big drawback of their method is that one of their factors require comparison of unpaired real and fake samples. Unfortunately, the generation of fake samples may not be possible if one cannot sample the conditionals (e.g. in the case of semantic maps).
\cite{tulyakov2017hybrid} solves the semi-supervised conditional generation problem, but for variational auto-encoders \cite{kingma2013vae}. Their main idea is to introduce a network \textit{q}, similar to our labeller network, which predicts conditionals of the data samples.
\section{Conclusion}
\label{sec:conclusion}
We have presented a semi-supervised framework \sscgan{} for training of cGANs with much fewer labels than traditionally required. Our approach is motivated by the observation that learning conditional mapping (which requires labelled data) is a much simpler problem in principle than learning complex distributions of real world datasets (which requires only unsupervised data). Thus, we reason that it should be possible to train a network for mapping conditionals with very few labelled training data. Our method introduces an additional labeller network which is jointly trained with the cGANs. We also propose an unsupervised GAN objective, which combined with the supervised objective achieves the dual purpose of both learning the underlying distribution, as well as learning the conditional mapping. Our method is very simple conceptually, as well as easy to implement. Moreover, it is very general allowing it to work for any cGAN setting.

\section{Acknowledgements}
\label{sec:acknowledgements}
We would like to thank Mohit Jain for help in setting up and running the AMT study, and B. Ashok for helpful discussions.

\clearpage

\bibliographystyle{plain}
\bibliography{references}

\clearpage

\appendix

\section{Semantic Image Synthesis}

\begin{figure*}[h]
  \centering
  \vspace{-28pt}
  \begin{tabular}{ccc}
    \includegraphics[width=.28\textwidth]{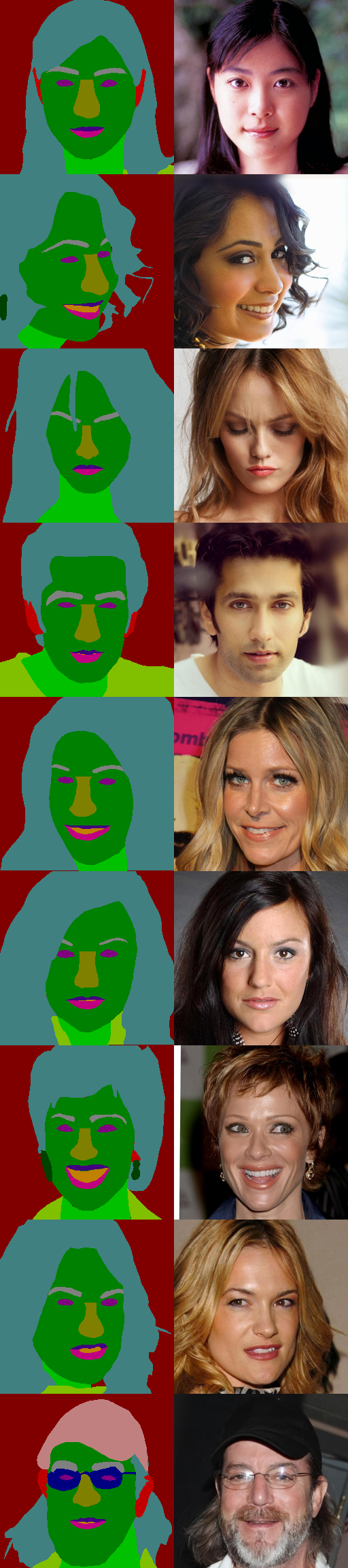} &  
    \includegraphics[width=.14\textwidth]{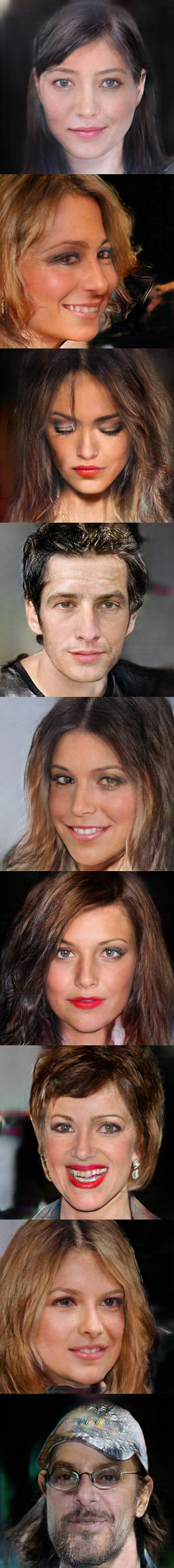}   &
    \includegraphics[width=.28\textwidth]{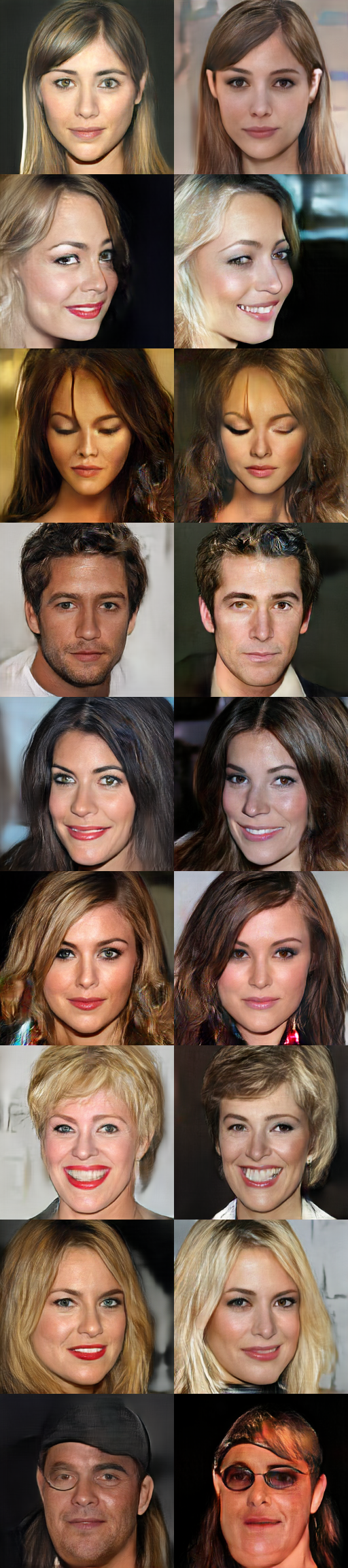} \\
    {\scriptsize Label, Ground Truth} &
    {\scriptsize Fully Supervised} &
    {\scriptsize \sscgan{}: 5, 25 labelled pairs}
  \end{tabular}
  \vspace{-2pt}
  \caption{Synthesis results (randomly chosen) for \camhq{} dataset on test set semantic maps. Shown are results from fully-supervised SPADE and \sscgan{} with 5 and 25 labelled pairs. \sscgan{} produces realistic results, while accurately respecting the semantic layout (even when trained with just 5 supervised pairs!).}
  \label{fig:camhq_rand_fully_sup_vs_ours}
\end{figure*}

\begin{figure*}[h]
  \centering
  \vspace{-28pt}
  \begin{tabular}{ccc}
    \includegraphics[width=.28\textwidth]{figures/camhq/random/random_grid_reals.png} &  
    \includegraphics[width=.28\textwidth]{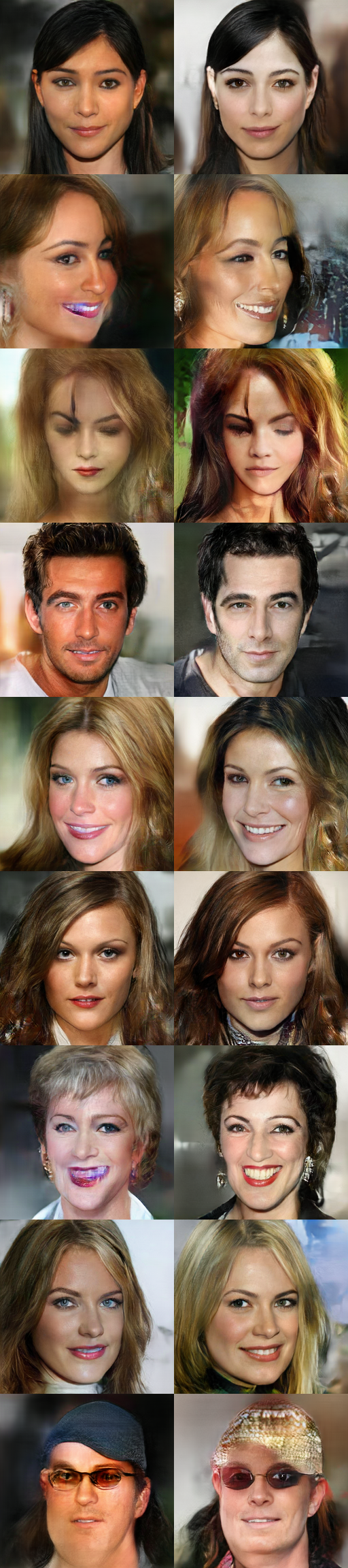}   &
    \includegraphics[width=.28\textwidth]{figures/camhq/random/random_grid_ours.png} \\
    {\scriptsize Label, Ground Truth} &
    {\scriptsize \bltwocap{}: 5, 25 labelled pairs} & {\scriptsize \sscgan{}: 5, 25 labelled pairs}
  \end{tabular}
  \vspace{-2pt}
  \caption{Comparison of \bltwocap{} and \sscgan{} synthesis results for \camhq{} test set on the same randomly chosen inputs as Figure~\ref{fig:camhq_rand_fully_sup_vs_ours}. Results are shown for training with 5 and 25 labelled pairs. \bltwocap{} struggles with synthesis quality, especially when labelled pairs are low, while \sscgan{} synthesis realistic images.}
  \label{fig:camhq_rand_bl2_vs_ours}
\end{figure*}

\begin{figure*}[t]
  \centering
  \resizebox{\textwidth}{!}{
  \begin{tabular}{cccc}
    \includegraphics[width=.4\textwidth]{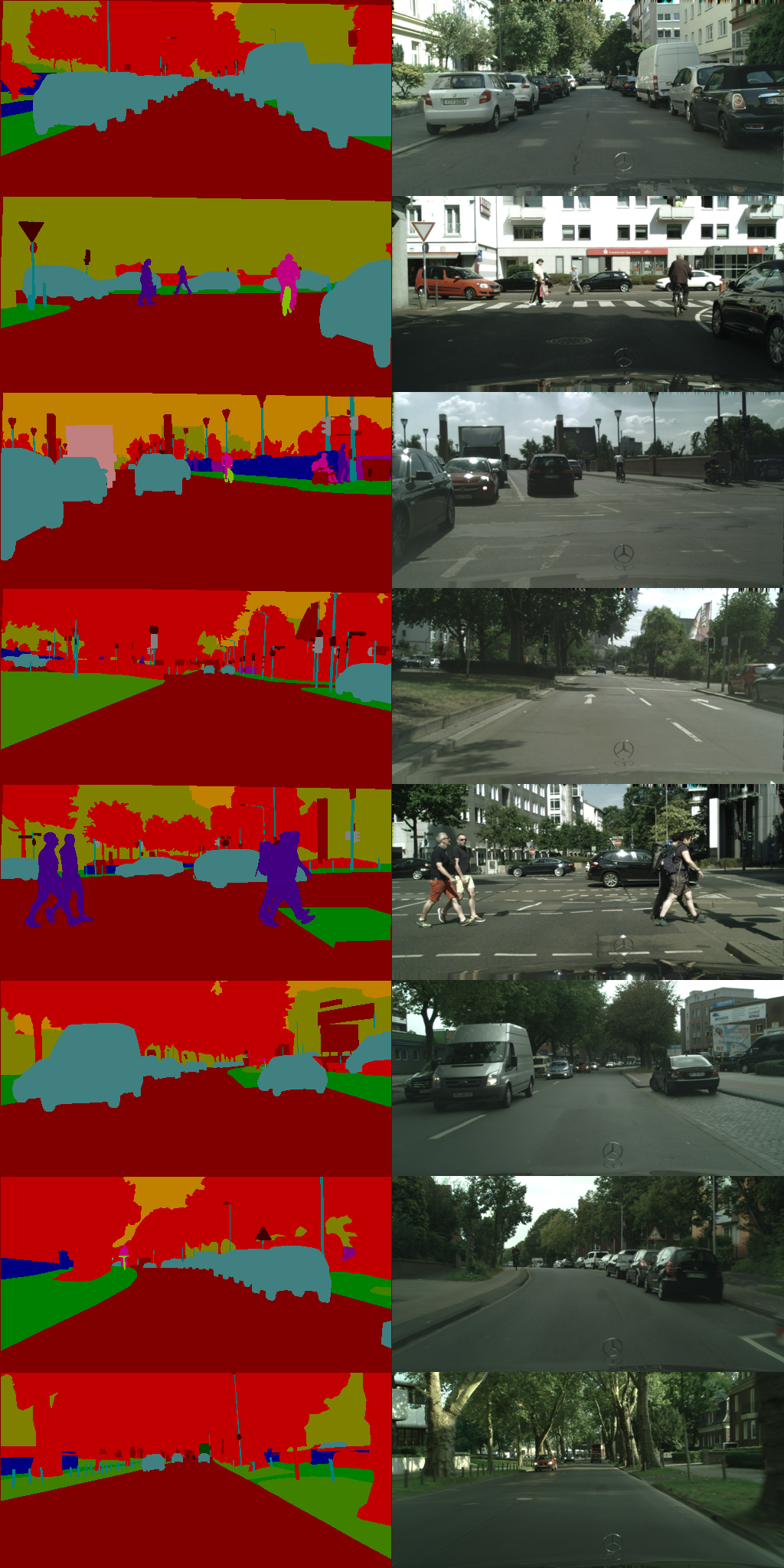} &
    \includegraphics[width=.2\textwidth]{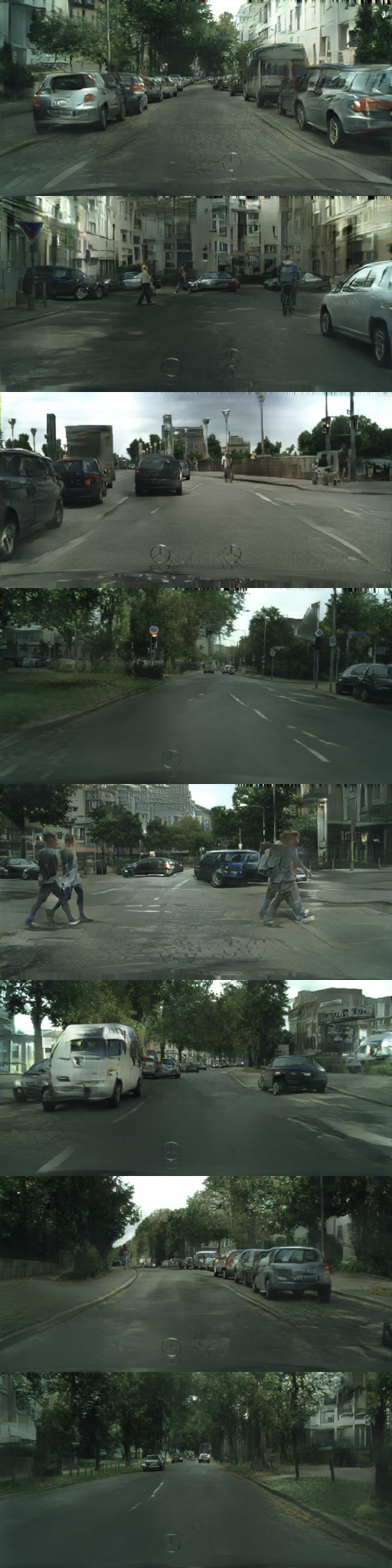}   &
    \includegraphics[width=.4\textwidth]{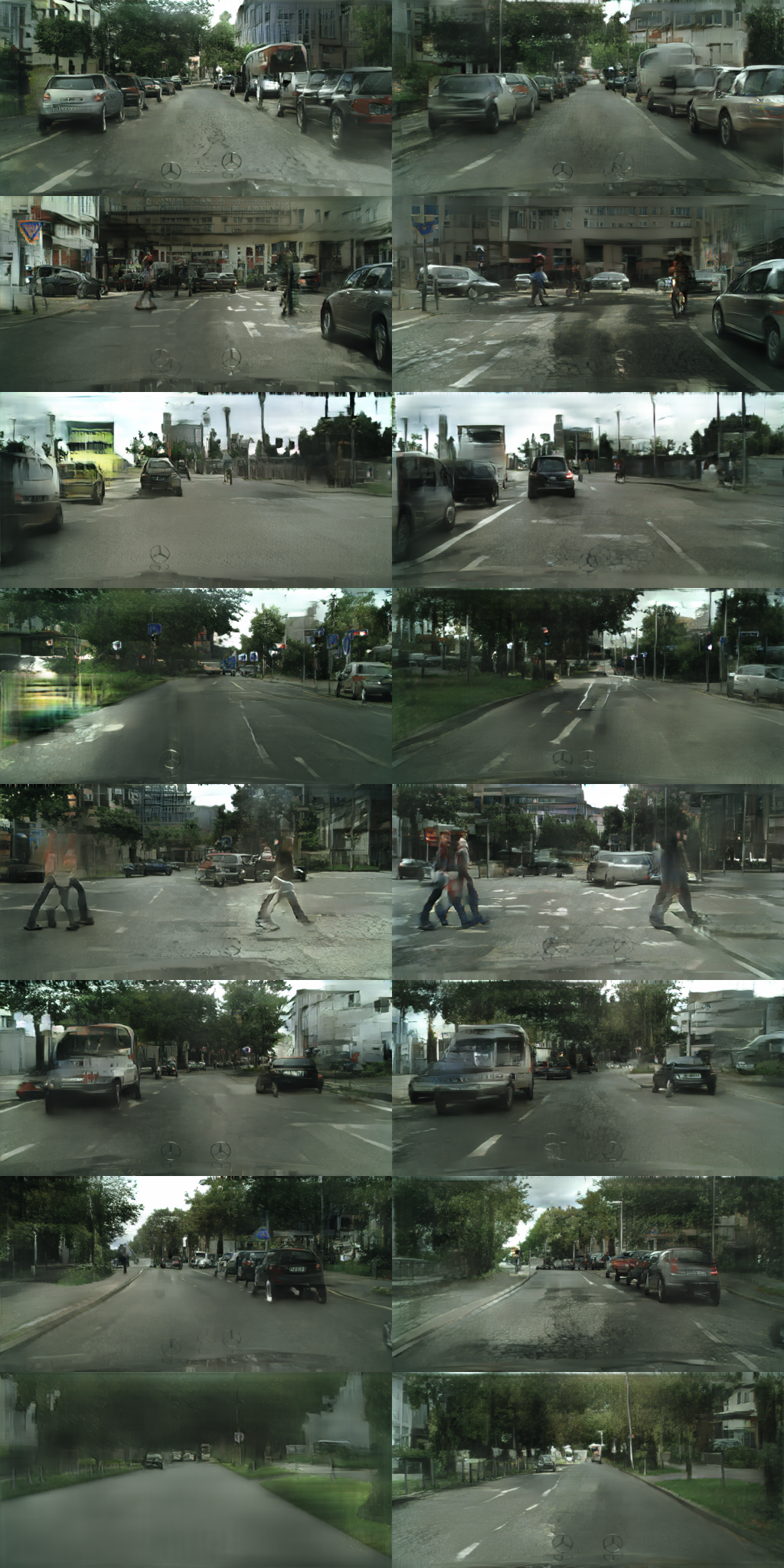} &
    \includegraphics[width=.4\textwidth]{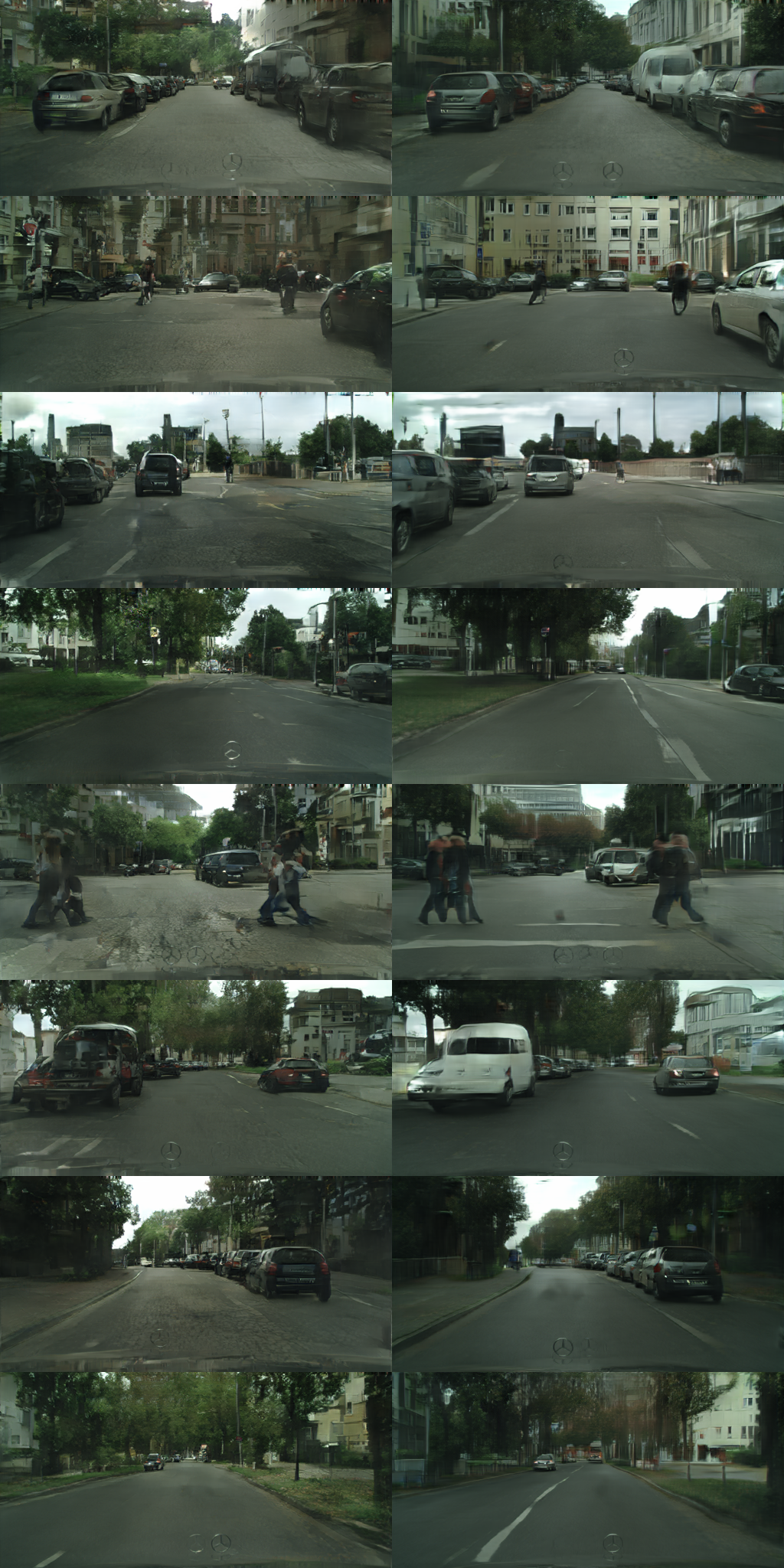} \\
  
   {\scriptsize Labels, Ground Truth} & {\scriptsize Fully Supervised} & {\scriptsize \bltwocap{}: 100, 500 labelled pairs} & {\scriptsize \sscgan{}: 100, 500 labelled pairs} \\
   & {\scriptsize FID: 58.76} & {\scriptsize FID: 73.71, 69.13} & {\scriptsize FID: 65.06, 53.83}
  
  \end{tabular}
  }
  
  \caption{Synthesis results (randomly chosen) for CityScapes dataset on semantic maps from the test set. Results are shown for fully-supervised SPADE, \bltwo{} and \sscgan{} along with the FID scores computed on the test set.}
  
  \label{fig:city_rand_results}
\end{figure*}

\medskip\noindent\textbf{More results:} We show more results for semantic image synthesis from our experiments. All results are from randomly chosen semantic maps from the test set. Figure~\ref{fig:camhq_rand_fully_sup_vs_ours} shows \sscgan{} results for experiments with 5 and 25 supervised pairs, compared to the fully-supervised (vanilla SPADE) baseline on the \camhq{} dataset. Comparison with the \bltwo{} is shown in figure~\ref{fig:camhq_rand_bl2_vs_ours}. Results with the CityScapes dataset are shown in  Figure~\ref{fig:city_rand_results}, which compares \sscgan{} to the baselines.

\begin{figure*}[t]
    \centering
    \begin{subfigure}[t]{1.0\textwidth}
        \centering
        \includegraphics[width=\textwidth]{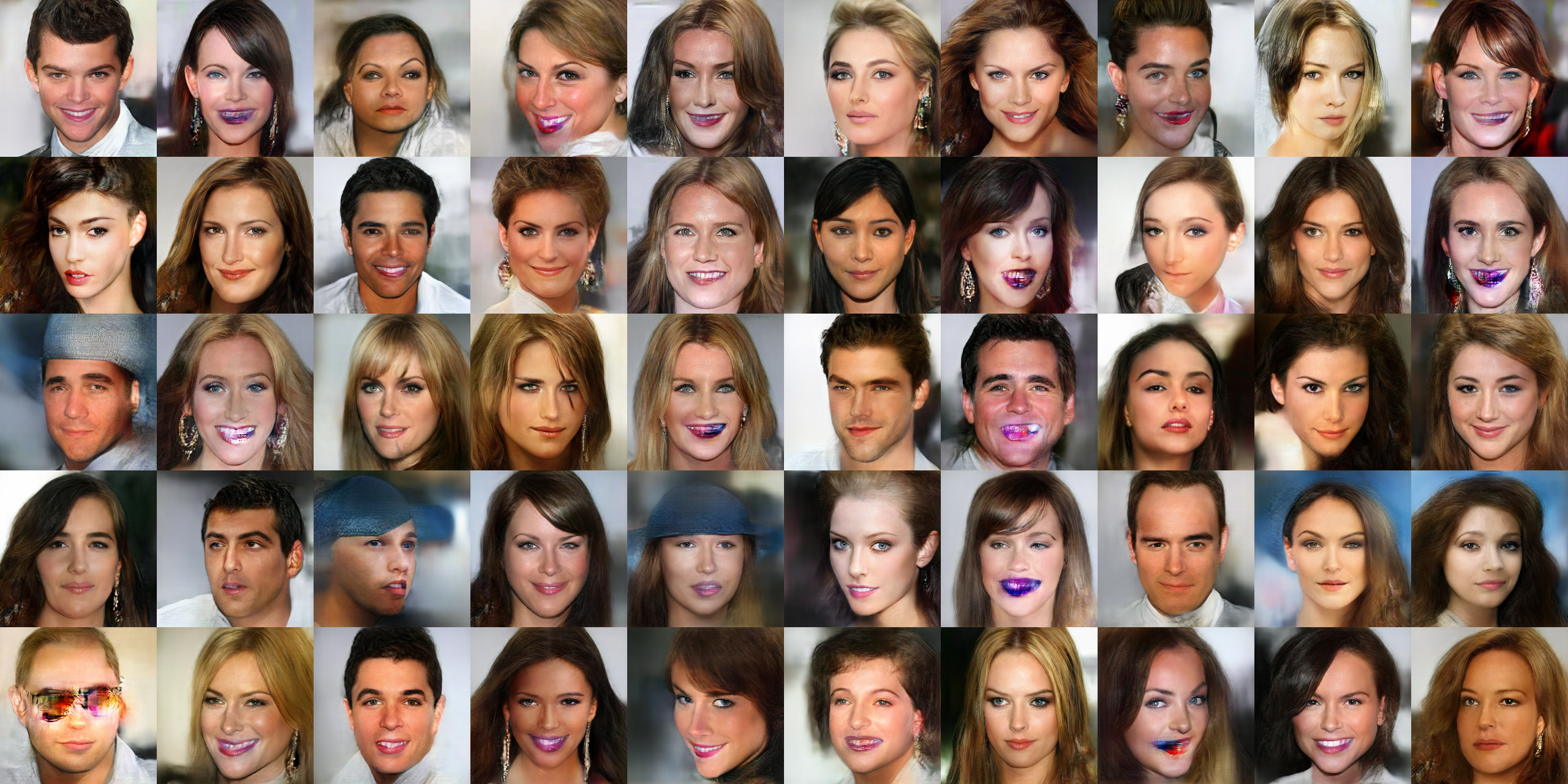}
        \caption{\bltwocap{} (5 supervised pairs). FID: 49.67}
    \end{subfigure}
    \par\medskip
    \begin{subfigure}[t]{1.0\textwidth}
        \centering
        \includegraphics[width=\textwidth]{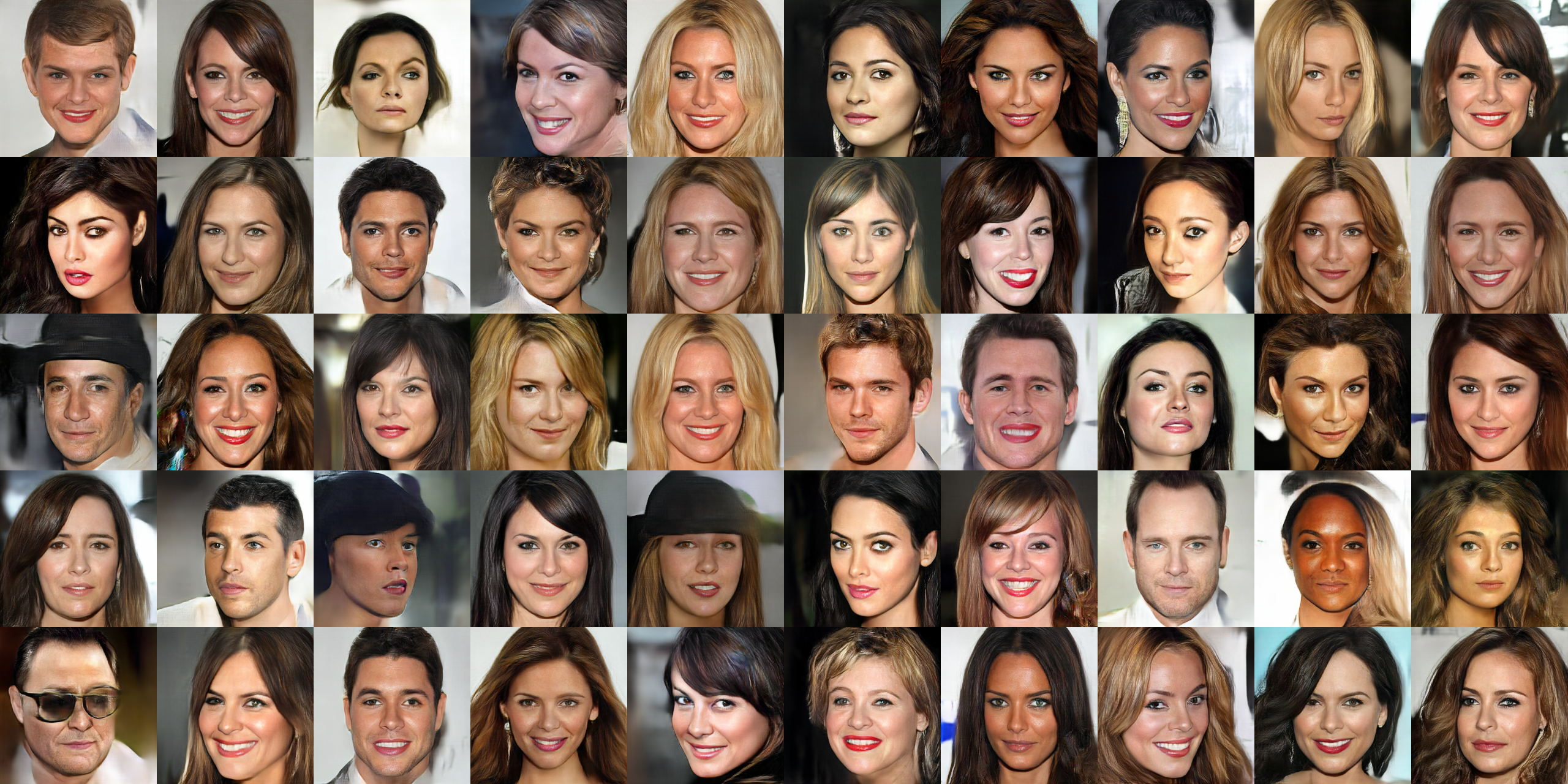}
        \caption{\sscgan{} (5 supervised pairs). FID: 50.83}
    \end{subfigure}
    \caption{Synthesis results and FID scores computed on the test set, for \camhq{} dataset with \bltwocap{} and \sscgan{} trained on 5 supervised pairs. The same semantic maps (randomly chosen from the test set) are used for both experiments. As can be seen, the FID scores in this case do not correlate with the visual quality of the results. The \bltwocap{} results have a lot of artifacts and look artificial, but still get a better (lower) FID score.}
    \label{fig:fid_camhq_5_bl2_vs_ours}
\end{figure*}

\medskip\noindent\textbf{FID score discrepancy:} As discussed in the paper, we found that FID scores did not correlate with visual quality of the synthesized results for \camhq{} dataset. This is illustrated in figure~\ref{fig:fid_camhq_5_bl2_vs_ours}, which shows synthesized results and the corresponding FID scores for the \bltwo{} and \sscgan{}, trained on 5 supervised pairs. Even though the visual quality of \bltwo{} is very bad and contains obvious artifacts, its FID score is superior to that of \sscgan{} which has much better visual quality results. The human perceptual study (Table 1 of main paper), however, yields results in line with the visual quality.

\begin{figure*}[t]
    \centering
    \includegraphics[width=0.9\textwidth]{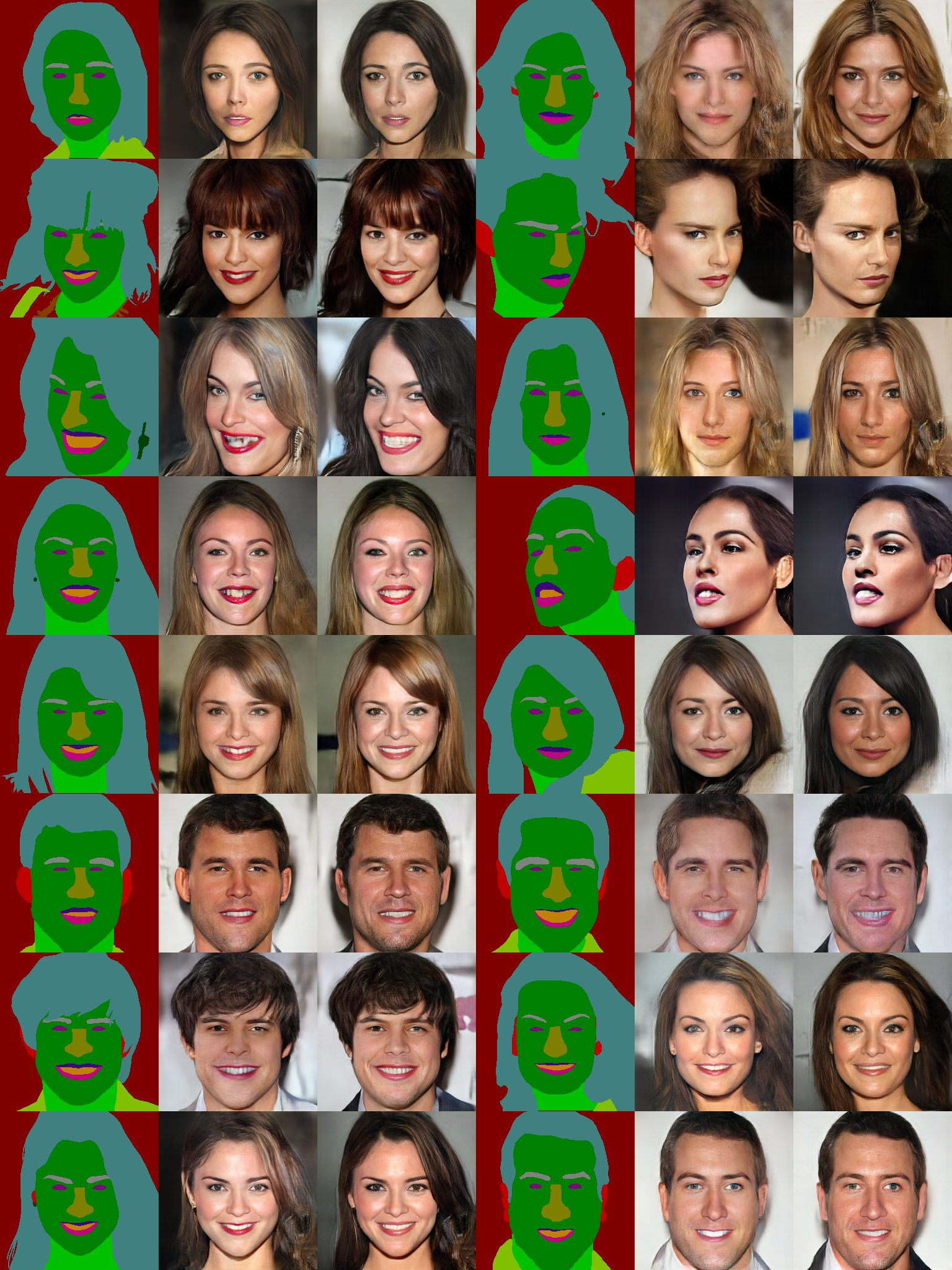}
    \caption{Shown are examples of one-pass inference vs two-pass inference for the \camhq{} dataset. The first (fourth) column shows input segementation maps, followed by one-pass results, followed by two-pass results. As shown, two-pass inference results in superior visual quality and less artifacts, compared to one-pass-inference.}
    \label{fig:camhq_onepass_vs_twopass}
\end{figure*}

\medskip\noindent\textbf{Two-pass Inference:} As mentioned in the paper, we do a two-pass inference to make sure that the input conditional is closer to the unsupervised distribution on which the \sscgan{} is mostly trained on. Figure~\ref{fig:camhq_onepass_vs_twopass} shows the results of standard one-pass inference along with the two-pass inference. As can be seen, the synthesis quality of two-pass inference is superior to that of one-pass inference.

\section{Class Conditional Synthesis}

Class conditional image synthesis looks at the problem of generating images for a given class input, such as dog, car, etc. For our evaluation, we incorporate the \sscgan{} framework into the BigGAN network \cite{brock2018biggan}. For the labeller network L, we used Resnet-18 classifier network \cite{resnet_he_2016}. Our integration required only $\approx200$ lines of code to add the requisite loss functions, etc. We plan to open-source our code which was built over a PyTorch implementation of BigGAN\footnote{https://github.com/ilyakava/BigGAN-PyTorch/}.

We evaluate on the Cifar-10 %
dataset which contains 10 classes with a total of 60,000 labelled images. The dataset is split into 50,000 training and 10,000 test images. For \sscgan{}, we used only a subset of the labeled training set to form our supervised set \Sset{}, while the remaining images are used without labels to form our unsupervised set \Uset

\medskip\noindent\textbf{Baselines:} Similar to section~\ref{sec:semantic_synthesis_expts}, our first baseline is the \textit{Fully supervised} baseline, i.e. vanilla BigGAN, where we train BigGAN with the full supervised training set. This baseline can be expected to give the best results as it uses the entire supervised training data. The second baseline is what we call the \textit{\bltwocap{}} baseline, where we first train the labeller network with the supervised subset \Sset{}, and use that to generate labels for the images in \Uset{}. BigGAN is then trained as usual with these synthetic labels as well as those of \Sset{}. We also compare against the S3GAN model proposed in \cite{lucic2019highfidfewerlabels}.

\medskip\noindent\textbf{Results:} We ran multiple experiments with different number of labelled examples (chosen randomly) for the supervised set. In particular we ran with 600 and 2600 pairs. Table~\ref{tab:biggan_cifar_scores} shows the test Inception score (IS) %
and FID score \cite{heusel2017ganstturfid} for the various experiments. As shown we perform better than the \bltwo{} in both metrics, and comparably to S3GAN. We would like to note that the S3GAN approach is only applicable for class conditional tasks (since it is based on the property of the discriminator being a classifier), and cannot be applied to tasks such as semantic map conditional synthesis. On the other hand \sscgan{} is a generic framework and can be applied to any conditional synthesis task.

\begin{table}
\resizebox{0.475\textwidth}{!}{
\begin{tabular}{cccc}
    \toprule
    Pairs & Method &
    \begin{tabular} 
        {c}IS \\ Best / Final 
    \end{tabular} &
    \begin{tabular} 
        {@{}c@{}}FID \\ Best / Final 
    \end{tabular} \\
    
    \midrule
    
    50,000 pairs & Fully Supervised & 9.18 / 9.18 & 11.07 / 11.07  \\
    
    \midrule

    600 pairs & 
    \begin{tabular} 
        {c}\bltwocapshort{} \\ \sscgan{} \\ S3GAN \\
    \end{tabular} &
    
    \begin{tabular} 
        {c}8.63 / 8.55 \\ 8.75 / 8.58 \\ 8.63 / 8.43 \\
    \end{tabular} &
    
    \begin{tabular}
        {c}16.54 / 16.54 \\ 14.30 / 14.30 \\ 14.50 / 14.82 \\
    \end{tabular} \\
    \midrule

    2600 pairs & 
    \begin{tabular} 
        {c}\bltwocapshort{} \\ \sscgan{} \\ S3GAN \\
    \end{tabular} &
    
    \begin{tabular} 
        {c}8.94 / 8.81 \\ 9.00 / 8.86 \\ 8.95 / 8.91 \\
    \end{tabular} &
    
    \begin{tabular}
        {c}13.49 / 13.49 \\ 12.02 / 12.62 \\ 11.92 / 12.48 \\
    \end{tabular} \\
    \bottomrule

\end{tabular}}

\caption{IS and FID metrics on the test split for Cifar-10. We report the best scores during training and the final scores. For FID, a lower score is better}
\label{tab:biggan_cifar_scores}
\end{table}

\begin{figure}[h]
  \centering
  \begin{subfigure}{.44\textwidth}
  \centering
    \includegraphics[width=1.0\textwidth]{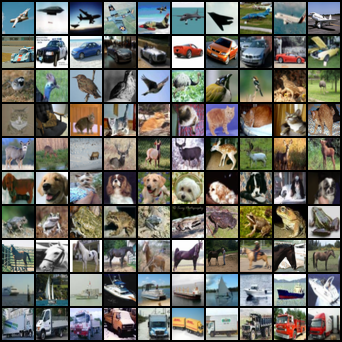}
    \caption{Reals}
  \end{subfigure} 
  \hspace{6pt}
  \begin{subfigure}{.44\textwidth}
  \centering
    \includegraphics[width=1.0\textwidth]{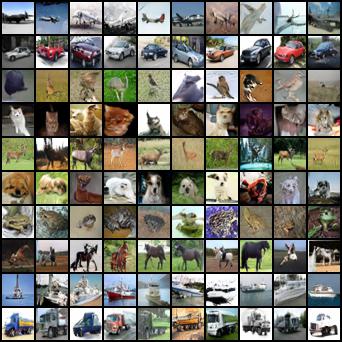}  
        \caption{Fully Supervised}
  \end{subfigure}
  \caption{Real images and class conditional synthesis results (randomly chosen) for the fully-supervised experiment (vanilla BigGAN) are shown for the Cifar-10 dataset.}
  \label{fig:cifar_reals_bl1}
\end{figure}

\begin{figure}[h]
  \centering
  
  \begin{subfigure}{.44\textwidth}
  \centering
    \includegraphics[width=1.0\textwidth]{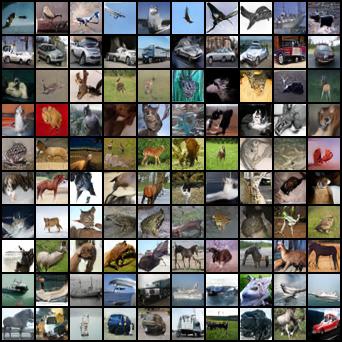}  
    \caption{\bltwocap{} (600 labels)}
  \end{subfigure}  
  \hspace{6pt}
  \begin{subfigure}{.44\textwidth}
  \centering
    \includegraphics[width=1.0\textwidth]{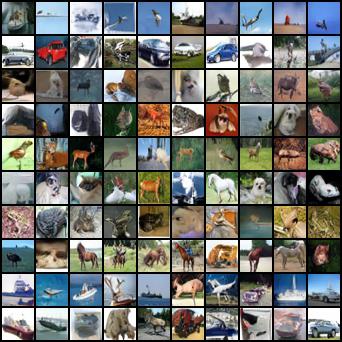}
        \caption{\sscgan{}{} (600 labels)}
  \end{subfigure}

  \vspace{-5pt}
    \caption{Class conditional synthesis results (randomly chosen) for Cifar-10 dataset. Results are shown for \bltwocap{} and \sscgan{} networks trained using 600 labels.}
    \label{fig:cifar_bl2_ours_600}
\end{figure}

\begin{figure}[h]
  \centering

  \begin{subfigure}{.44\textwidth}
  \centering
    \includegraphics[width=1.0\textwidth]{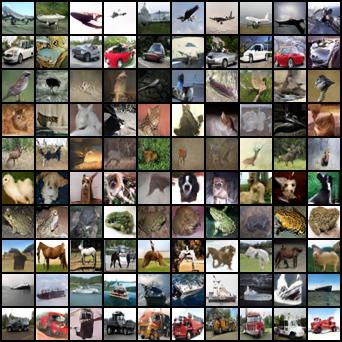}  
    \caption{\bltwocap{} (2600 labels)}
  \end{subfigure}  
  \hspace{6pt}
  
  \begin{subfigure}{.44\textwidth}
  \centering
    \includegraphics[width=1.0\textwidth]{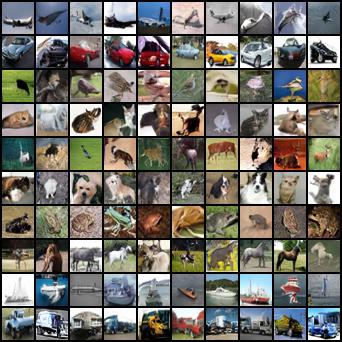}
        \caption{\sscgan{}{} (2600 labels)}
  \end{subfigure}
  \vspace{-5pt}
    \caption{Class conditional synthesis results (randomly chosen) for Cifar-10 dataset. Results are shown for \bltwocap{} and \sscgan{} networks trained using 2600 labels.}
    \label{fig:cifar_bl2_ours_2600}
\end{figure}

We show results for class conditional synthesis on the Cifar-10 dataset. Figure~\ref{fig:cifar_reals_bl1} shows randomly chosen real images, as well as fake images from the fully supervised training (vanilla BigGAN) experiment. Each row correspond to one class, in the following order from top to bottom: airplane, automobile, bird, cat, deer, dog, frog, horse, ship, truck. Figure~\ref{fig:cifar_bl2_ours_600},~\ref{fig:cifar_bl2_ours_2600} show randomly chosen images from \bltwo{} and \sscgan{} with 600 and 2.6k training pairs, respectively.

\end{document}